%% file: main.tex
\documentclass{article}
\usepackage{selectp}

\usepackage{microtype}
\usepackage{graphicx}
\usepackage{subfigure}
\usepackage{booktabs} %

\usepackage{hyperref}
\usepackage{url}
\usepackage{caption} %
\usepackage{floatrow}
\usepackage{fmtcount} %
\usepackage[titletoc,title]{appendix}
\usepackage{enumitem}
\usepackage{multirow}

\usepackage{floatrow} %
\newfloatcommand{capbtabbox}{table}[][\FBwidth]

\usepackage{wrapfig} %

\usepackage{fancyvrb} %

\usepackage{amsmath}
\usepackage{amssymb}
\usepackage{algorithm, algorithmic}

\usepackage{multirow}

\usepackage{pifont}%

\usepackage{hyperref}

\usepackage[accepted]{icml2020}

\definecolor{mygreen}{rgb}{0, 0.5, 0.0}

\newcommand{\Var}{\mathrm{Var}}
\newcommand{\valstd}[2]{$#1 {\scriptstyle \,\pm\, #2}$}

\icmltitlerunning{High-Performance Normalizer-Free ResNets}

\begin{document}

\twocolumn[

\icmltitle{High-Performance Large-Scale Image Recognition Without Normalization}

\icmlsetsymbol{equal}{*}

\begin{icmlauthorlist}
\icmlauthor{Andrew Brock}{dm}
\icmlauthor{Soham De}{dm}
\icmlauthor{Samuel L. Smith}{dm}
\icmlauthor{Karen Simonyan}{dm}

\end{icmlauthorlist}

\icmlaffiliation{dm}{DeepMind, London, United Kingdom}

\icmlcorrespondingauthor{Andrew Brock}{ajbrock@google.com}

\icmlkeywords{Machine Learning, ICML, normalization, NFNets, normalizer-free, ImageNet,}

\vskip 0.3in
]

\printAffiliationsAndNotice{}  %

\begin{abstract}
\setcounter{footnote}{1}
Batch normalization is a key component of most image classification models, but it has many undesirable properties stemming from its dependence on the batch size and interactions between examples.
Although recent work has succeeded in training deep ResNets without normalization layers, these models do not match the test accuracies of the best batch-normalized networks, and are often unstable for large learning rates or strong data augmentations.
In this work, we develop an adaptive gradient clipping technique which overcomes these instabilities, and design a significantly improved class of Normalizer-Free ResNets.
Our smaller models match the test accuracy of an EfficientNet-B7 on ImageNet while being up to $8.7\times$ faster to train, and our largest models attain a new state-of-the-art top-1 accuracy of $86.5\%$.
In addition, Normalizer-Free models attain significantly better performance than their batch-normalized counterparts when fine-tuning on ImageNet after large-scale pre-training on a dataset of 300 million labeled images, with our best models obtaining an accuracy of $89.2\%$.\footnote{Code available at \url{https://github.com/deepmind/deepmind-research/tree/master/nfnets}}

\end{abstract}

\section{Introduction}
\label{sec:intro}
\input{Section1_Intro.tex}

\input{Section2_Background.tex}

\section{Adaptive Gradient Clipping for Efficient Large-Batch Training}
\label{sec:UAGC}

\input{Section3_GradClipping.tex}

\section{Normalizer-Free Architectures with Improved Accuracy and Training Speed}
\label{sec:modeldesign}

\input{Section5_ModelDesign.tex}

\section{Experiments}
\label{sec:experiments}

\input{Section6_Experiments}

\section*{Conclusion}
\label{sec:conclusion}
\input{Section7_Conclusion.tex}

\section*{Acknowledgements}
We would like to thank A\"{a}ron van den Oord, Sander Dieleman, Erich Elsen, Guillaume Desjardins, Michael Figurnov, Nikolay Savinov, Omar Rivasplata, Relja Arandjelovi\'{c}, and Rishub Jain for helpful discussions and guidance. Additionally, we would like to thank Blake Hechtman, Tim Shen, Peter Hawkins, and James Bradbury for assistance with developing highly performant JAX code.

\bibliography{icml2020}
\bibliographystyle{icml2020}

\begin{appendices}

\clearpage
\section{Experiment Details} 
\label{appendix:experiment_details}
\input{Appendix_Experiment_Details.tex}

\clearpage
\section{Downsides of Batch Normalization}
\label{appendix:batchnorm_bad}
\input{Appendix_WhyBN_Bad.tex}

\clearpage
\section{Model Details} 
\label{appendix:model_details}

\input{Appendix_Model_Details.tex}

\clearpage
\section{Additional AGC Ablations}
\label{appendix:agc_ablations}
\input{Appendix_AGC_Ablations.tex}

\clearpage
\section{Negative Results}
\label{appendix:negative_results}
\input{Appendix_Negative_Results.tex}

\end{appendices}

\end{document}

%% file: Section1_Intro.tex
\begin{figure}[t]
	\includegraphics[width=0.99\columnwidth]{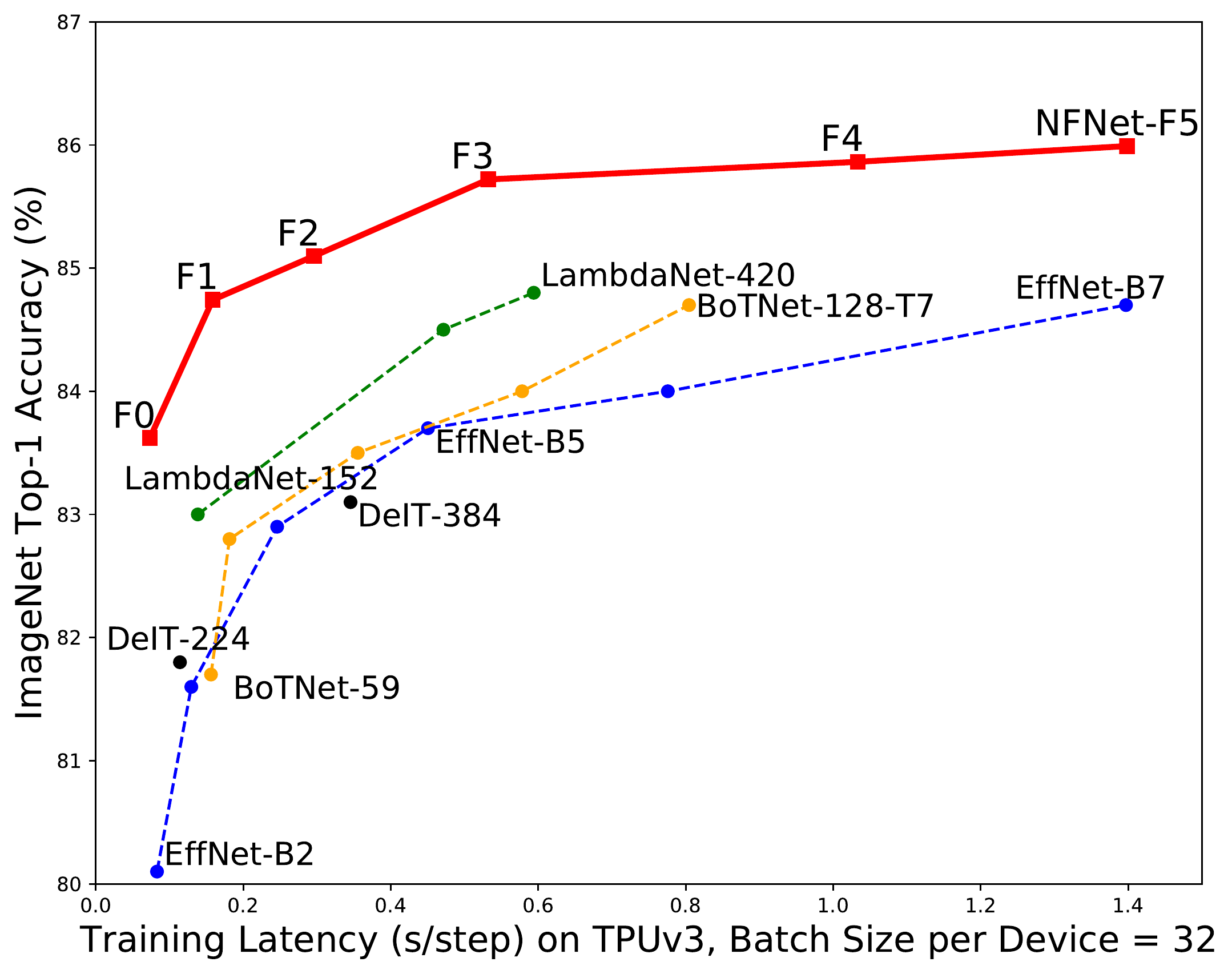}
\vskip -3mm
\caption{\label{fig:train_pareto_front}\textbf{ImageNet Validation Accuracy vs Training Latency.} All numbers are single-model, single crop. Our NFNet-F1 model achieves comparable accuracy to an EffNet-B7 while being 8.7$\times$ faster to train. Our NFNet-F5 model has similar training latency to EffNet-B7, but achieves a state-of-the-art 86.0\% top-1 accuracy on ImageNet. We further improve on this using Sharpness Aware Minimization \citep{foret2021sharpnessaware} to achieve 86.5\% top-1 accuracy.}
\vskip-2mm
\end{figure}

The vast majority of recent models in computer vision are variants of deep residual networks \citep{he2016resnets, he2016identity}, trained with batch normalization \citep{ioffe2015batchnorm}. The combination of these two architectural innovations has enabled practitioners to train significantly deeper networks which can achieve higher accuracies on both the training set and the test set. Batch normalization also smoothens the loss landscape \citep{santurkar2018does}, which enables stable training with larger learning rates and at larger batch sizes \citep{bjorck2018understanding, de2020batch}, and it can have a regularizing effect \citep{hoffer2017train,luo2018towards}.

However, batch normalization has three significant practical disadvantages. First, it is a surprisingly expensive computational primitive, which incurs memory overhead \citep{rota2018place}, and significantly increases the time required to evaluate the gradient in some networks \citep{gitman2017comparison}. 
Second, it introduces a discrepancy between the behaviour of the model during training and at inference time \citep{summers2019four, singh2019evalnorm}, introducing hidden hyper-parameters that have to be tuned. Third, and most importantly, batch normalization breaks the independence between training examples in the minibatch. 

This third property has a range of negative consequences. For instance, practitioners have found that batch normalized networks are often difficult to replicate precisely on different hardware, and batch normalization is often the cause of subtle implementation errors, especially during distributed training \citep{pham2019cradle}. Furthermore, batch normalization cannot be used for some tasks, since the interaction between training examples in a batch enables the network to `cheat' certain loss functions. For example, batch normalization requires specific care to prevent information leakage in some contrastive learning algorithms \citep{chen2020simple, he2020momentum}. This is a major concern for sequence modeling tasks as well, which has driven language models to adopt alternative normalizers \citep{ba2016layer, vaswani2017attention}. 
The performance of batch-normalized networks can also degrade if the batch statistics have a large variance during training \citep{shen2020powernorm}.
Finally, the performance of batch normalization is sensitive to the batch size, and batch normalized networks perform poorly when the batch size is too small \citep{hoffer2017train, ioffe2017batch, wu2018group}, which limits the maximum model size we can train on finite hardware. We expand on the challenges associated with batch normalization in Appendix \ref{appendix:batchnorm_bad}.

Therefore, although batch normalization has enabled the deep learning community to make substantial gains in recent years, we anticipate that in the long term it is likely to impede progress. We believe the community should seek to identify a simple alternative which achieves competitive test accuracies and can be used for a wide range of tasks.
Although a number of alternative normalizers have been proposed \citep{ba2016layer, wu2018group, huang2020normalization}, these alternatives often achieve inferior test accuracies and introduce their own disadvantages, such as additional compute costs at inference. Fortunately, in recent years two promising research themes have emerged. The first studies the origin of the benefits of batch normalization during training \citep{balduzzi2017shattered, santurkar2018does, bjorck2018understanding, luo2018towards, yang2019mean, jacot2019freeze, de2020batch}, while the second seeks to train deep ResNets to competitive accuracies without normalization layers \citep{hanin2018start, zhang2019fixup, de2020batch, shao2020normalization, brock2021characterizing}.

A key theme in many of these works is that it is possible to train very deep ResNets without normalization by suppressing the scale of the hidden activations on the residual branch.
The simplest way to achieve this is to introduce a learnable scalar at the end of each residual branch, initialized to zero \citep{goyal2017accurate, zhang2019fixup, de2020batch, bachlechner2020rezero}. However this trick alone is not sufficient to obtain competitive test accuracies on challenging benchmarks. Another line of work has shown that ReLU activations introduce a `mean shift', which causes the hidden activations of different training examples to become increasingly correlated as the network depth increases \citep{huang2017centered, jacot2019freeze}. In a recent work, \citet{brock2021characterizing} introduced ``Normalizer-Free'' ResNets, which suppress the residual branch at initialization and apply Scaled Weight Standardization \citep{qiao2019weight} to remove the mean shift. With additional regularization, these unnormalized networks match the performance of batch-normalized ResNets \citep{he2016identity} on ImageNet \citep{ILSVRC2015}, but they are not stable at large batch sizes and do not match the performance of EfficientNets \citep{tan2019efficientnet}, the current state of the art \citep{gong2020maxup}. This paper builds on this line of work and seeks to address these central limitations. Our main contributions are as follows:

\begin{itemize}

    \item %
    We propose Adaptive Gradient Clipping (AGC), which clips gradients based on the \textit{unit-wise ratio of gradient norms to parameter norms}, and we demonstrate that AGC allows us to train Normalizer-Free Networks with larger batch sizes and stronger data augmentations.
    
    \item We design a family of Normalizer-Free ResNets, called NFNets, which set new state-of-the-art validation accuracies on ImageNet for a range of training latencies (See Figure~\ref{fig:train_pareto_front}). 
    Our NFNet-F1 model achieves similar accuracy to EfficientNet-B7 while being 8.7$\times$ faster to train, and
    our largest model sets a new overall state of the art without extra data of $86.5\%$ top-1 accuracy.

    \item We show that NFNets achieve substantially higher validation accuracies than batch-normalized networks when fine-tuning on ImageNet after pre-training on a large private dataset of 300 million labelled images. Our best model achieves $89.2\%$ top-1 after fine-tuning. %

\end{itemize}
    
The paper is structured as follows. We discuss the benefits of batch normalization in Section~\ref{subsec:batchnorm_benefits}, and recent work seeking to train ResNets without normalization in Section~\ref{sec:removing_batchnorm}. We introduce AGC in Section~\ref{sec:UAGC}, and we describe how we developed our new state-of-the-art architectures in Section~\ref{sec:modeldesign}. Finally, we present our experimental results in Section~\ref{sec:experiments}.

%% file: Section2_Background.tex
\section{Understanding Batch Normalization}
\label{subsec:batchnorm_benefits}
In order to train networks without normalization to competitive accuracy, we must understand the benefits batch normalization brings during training, and identify alternative strategies to recover these benefits. Here we list the four main benefits which have been identified by prior work.

\textbf{Batch normalization downscales the residual branch:} The combination of skip connections \citep{srivastava2015highway, he2016resnets, he2016identity} and batch normalization \citep{ioffe2015batchnorm} enables us to train significantly deeper networks with thousands of layers \citep{zhang2019fixup}. This benefit arises because batch normalization, when placed on the residual branch (as is typical), reduces the scale of hidden activations on the residual branches at initialization \citep{de2020batch}. This biases the signal towards the skip path, which ensures that the network has well-behaved gradients early in training, enabling efficient optimization \citep{balduzzi2017shattered, hanin2018start, yang2019mean}.

\textbf{Batch normalization eliminates mean-shift:} Activation functions like ReLUs or GELUs \citep{hendrycks2016gaussian}, which are not anti-symmetric, have non-zero mean activations. Consequently, the inner product between the activations of independent training examples immediately after the non-linearity is typically large and positive, even if the inner product between the input features is close to zero. This issue compounds as the network depth increases, and introduces a `mean-shift' in the activations of different training examples on any single channel proportional to the network depth \citep{de2020batch}, which can cause deep networks to predict the same label for all training examples at initialization \citep{jacot2019freeze}. Batch normalization ensures the mean activation on each channel is zero across the current batch, eliminating mean shift \citep{brock2021characterizing}.

\textbf{Batch normalization has a regularizing effect:} It is widely believed that batch normalization also acts as a regularizer enhancing test set accuracy, due to the noise in the batch statistics which are computed on a subset of the training data \citep{luo2018towards}. Consistent with this perspective, the test accuracy of batch-normalized networks can often be improved by tuning the batch size, or by using ghost batch normalization in distributed training \citep{hoffer2017train}.

\textbf{Batch normalization allows efficient large-batch training:} Batch normalization smoothens the loss landscape \citep{santurkar2018does}, and this increases the largest stable learning rate \citep{bjorck2018understanding}. While this property does not have practical benefits when the batch size is small \citep{de2020batch}, the ability to train at larger learning rates is essential if one wishes to train efficiently with large batch sizes. Although large-batch training does not achieve higher test accuracies within a fixed epoch budget \citep{smith2020generalization}, it does achieve a given test accuracy in fewer parameter updates, significantly improving training speed when parallelized across multiple devices \citep{goyal2017accurate}.

\section{Towards Removing Batch Normalization}
\label{sec:removing_batchnorm}

Many authors have attempted to train deep ResNets to competitive accuracies without normalization, by recovering one or more of the benefits of batch normalization described above. Most of these works suppress the scale of the activations on the residual branch at initialization, by introducing either small constants or learnable scalars \citep{hanin2018start, zhang2019fixup, de2020batch, shao2020normalization}. Additionally, \citet{zhang2019fixup} and \citet{de2020batch} observed that the performance of unnormalized ResNets can be improved with additional regularization. However only recovering these two benefits of batch normalization is not sufficient to achieve competitive test accuracies on challenging benchmarks \citep{de2020batch}.

In this work, we adopt and build on ``Normalizer-Free ResNets'' (NF-ResNets) \citep{brock2021characterizing}, %
a class of pre-activation ResNets \citep{he2016identity} which can be trained
to competitive training and test accuracies without normalization layers. NF-ResNets employ a residual block of the form $h_{i+1} = h_i + \alpha f_i(h_i/\beta_i)$, where $h_i$ denotes the inputs to the $i^{th}$ residual block, and $f_i$ denotes the function computed by the $i^{th}$ residual branch. The function $f_i$ is parameterized to be variance preserving at initialization, such that $\text{Var}(f_i(z)) = \text{Var}(z)$ for all $i$. The scalar $\alpha$ specifies the rate at which the variance of the activations increases after each residual block (at initialization), and is typically set to a small value like $\alpha = 0.2$. The scalar $\beta_i$ is determined by predicting the standard deviation of the inputs to the $i^{th}$ residual block, $\beta_i = \sqrt{\text{Var}(h_i)}$, where $\text{Var}(h_{i+1}) = \text{Var}(h_{i}) + \alpha^2$, except for transition blocks (where spatial downsampling occurs),
for which the skip path operates on the downscaled input $(h_i/\beta_i)$, and the expected variance is reset after the transition block to $h_{i+1} = 1 + \alpha^2$. 
The outputs of squeeze-excite layers \citep{hu2018squeeze} are multiplied by a factor of 2. Empirically,  \citet{brock2021characterizing} found it was also beneficial to include a learnable scalar initialized to zero at the end of each residual branch (`SkipInit' \citep{de2020batch}).

In addition, \citet{brock2021characterizing} prevent the emergence of a mean-shift in the hidden activations by introducing Scaled Weight Standardization (a minor modification of Weight Standardization \citep{huang2017centered, qiao2019weight}). This technique reparameterizes the convolutional layers as:
\begin{equation}
    \hat{W_{ij}} = \frac{W_{ij} - \mu_i }{\sqrt{N} \sigma_i },
    \label{eq:scaled_ws}
\end{equation}
where $\mu_i = (1/N) \sum_j W_{ij}$, $\sigma_i^2 = (1/N) \sum_j (W_{ij} - \mu_i)^2$, and $N$ denotes the fan-in. The activation functions are also scaled by a non-linearity specific scalar gain $\gamma$, which ensures that the combination of the $\gamma$-scaled activation function and a Scaled Weight Standardized layer is variance preserving. For ReLUs, $\gamma = \sqrt{2/(1- (1/\pi))}$ \citep{arpit2016normalization}. We refer the reader to \citet{brock2021characterizing} for a description of how to compute $\gamma$ for other non-linearities. 

With additional regularization (Dropout \citep{srivastava2014dropout} and Stochastic Depth \citep{huang2016deep}), Normalizer-Free ResNets match the test accuracies achieved by batch normalized pre-activation ResNets on ImageNet at batch size 1024. They also significantly outperform their batch normalized counterparts when the batch size is very small, but they perform worse than batch normalized networks for large batch sizes (4096 or higher). Crucially, they do not match the performance of state-of-the-art networks like EfficientNets \citep{tan2019efficientnet, gong2020maxup}.

%% file: Section3_GradClipping.tex
To scale NF-ResNets to larger batch sizes, we explore a range of gradient clipping strategies \citep{pascanu2013difficulty}. Gradient clipping is often used in language modeling to stabilize training \citep{merity2018regularizing}, and recent work shows that it allows training with larger learning rates compared to gradient descent, accelerating convergence \citep{zhang2019gradient}. This is particularly important for poorly conditioned loss landscapes or when training with large batch sizes, since in these settings the optimal learning rate is constrained by the maximum stable learning rate \citep{smith2020generalization}. We therefore hypothesize that gradient clipping should help scale NF-ResNets efficiently to the large-batch setting.

Gradient clipping is typically performed by constraining the norm of the gradient \citep{pascanu2013difficulty}. Specifically, for gradient vector $G = \partial{L}/\partial{\theta}$, where $L$ denotes the loss and $\theta$ denotes a vector with all model parameters, the standard clipping algorithm clips the gradient before updating $\theta$ as:
\begin{equation}
G \rightarrow
    \begin{cases}
    
    \lambda \frac{G}{\|G\|}& \text{if $\|G\| > \lambda$}, \\
    G & \text{otherwise.}
    \end{cases}
\end{equation}
The clipping threshold $\lambda$ is a hyper-parameter which must be tuned. Empirically, we found that while this clipping algorithm enabled us to train at higher batch sizes than before, training stability was extremely sensitive to the choice of the clipping threshold, requiring fine-grained tuning when varying the model depth, the batch size, or the learning rate.

To overcome this issue, %
we introduce ``Adaptive Gradient Clipping" (AGC), which we now describe. Let $W^\ell\in \mathbb{R}^{N \times M}$ denote the weight matrix of the $\ell^{th}$ layer, $G^\ell\in \mathbb{R}^{N \times M}$ denote the gradient with respect to $W^\ell$, and $\| \cdot \|_F$ denote the Frobenius norm, i.e., $\|W^{\ell}\|_F = \sqrt{\sum_i^N{\sum_j^M{(W^{\ell}_{i,j})^2}}}$. The AGC algorithm is motivated by the observation that the ratio of the norm of the gradients $G^{\ell}$ to the norm of the weights $W^\ell$ of layer $\ell$, $\frac{\|G^{\ell}\|_F}{\|W^{\ell}\|_F}$, provides a simple measure of how much a single gradient descent step will change  the original weights $W^{\ell}$. For instance, if we train using gradient descent without momentum, then $\frac{\|\Delta W^{\ell}\|}{\|W^{\ell}\|} =  h \frac{\|G^{\ell}\|_F}{\|W^{\ell}\|_F}$, where the parameter update for the $\ell^{th}$ layer is given by $\Delta W^{\ell} = -h G^{\ell}$, and $h$ is the learning rate. 

Intuitively, we expect training to become unstable if $(\|\Delta W^{\ell}\|/\|W^{\ell}\|)$ is large, which motivates a clipping strategy based on the ratio $\frac{\|G^{\ell}\|_F}{\|W^{\ell}\|_F}$. %
However in practice, we clip gradients based on the \textit{unit-wise ratios} of gradient norms to parameter norms, which we found to  perform better empirically than taking layer-wise norm ratios. Specifically, in our AGC algorithm, each unit $i$ of the gradient of the $\ell$-th layer $G_i^\ell$ (defined as the $i^{th}$ row of matrix $G^\ell$) is clipped as:
\begin{equation}
G^{\ell}_i \rightarrow
    \begin{cases}
    
    \lambda \frac{\|W^{\ell}_i\|^\star_F}{\|G^{\ell}_i\|_F}G^{\ell}_i& \text{if $\frac{\|G^{\ell}_i\|_F}{\|W^{\ell}_i\|^\star_F} > \lambda$}, \\
    G^{\ell}_i & \text{otherwise.}
    \end{cases}
\end{equation}
The clipping threshold $\lambda$ is a scalar hyperparameter, and we define $\|W_i\|^\star_F = \max(\|W_i\|_F, \epsilon)$, with default $\epsilon = 10^{-3}$, which prevents zero-initialized parameters from always having their gradients clipped to zero. For parameters in convolutional filters, we evaluate the unit-wise norms over the fan-in extent (including the channel and spatial dimensions). Using AGC, we can train NF-ResNets stably with larger batch sizes (up to 4096), as well as with very strong data augmentations like RandAugment \citep{cubuk2020randaugment} for which NF-ResNets without AGC fail to train \citep{brock2021characterizing}. Note that the optimal clipping parameter $\lambda$ may depend on the choice of optimizer, learning rate and batch size. Empirically, we find $\lambda$ should be smaller for larger batches.

AGC is closely related to a recent line of work studying ``normalized optimizers" \citep{you2017large, bernstein2020distance, you2019large}, which ignore the scale of the gradient by choosing an adaptive learning rate inversely proportional to the gradient norm. In particular, \citet{you2017large} propose LARS, a momentum variant which sets the norm of the parameter update to be a fixed ratio of the parameter norm, completely ignoring the gradient magnitude. AGC can be interpreted as a relaxation of normalized optimizers, which imposes a maximum update size based on the parameter norm but does not simultaneously impose a lower-bound on the update size or ignore the gradient magnitude. %
Although we are also able to stably train at high batch sizes with LARS, we found that doing so degrades performance.

\begin{figure}[t]
\subfigure[]{
	\includegraphics[width=0.48\columnwidth]
	{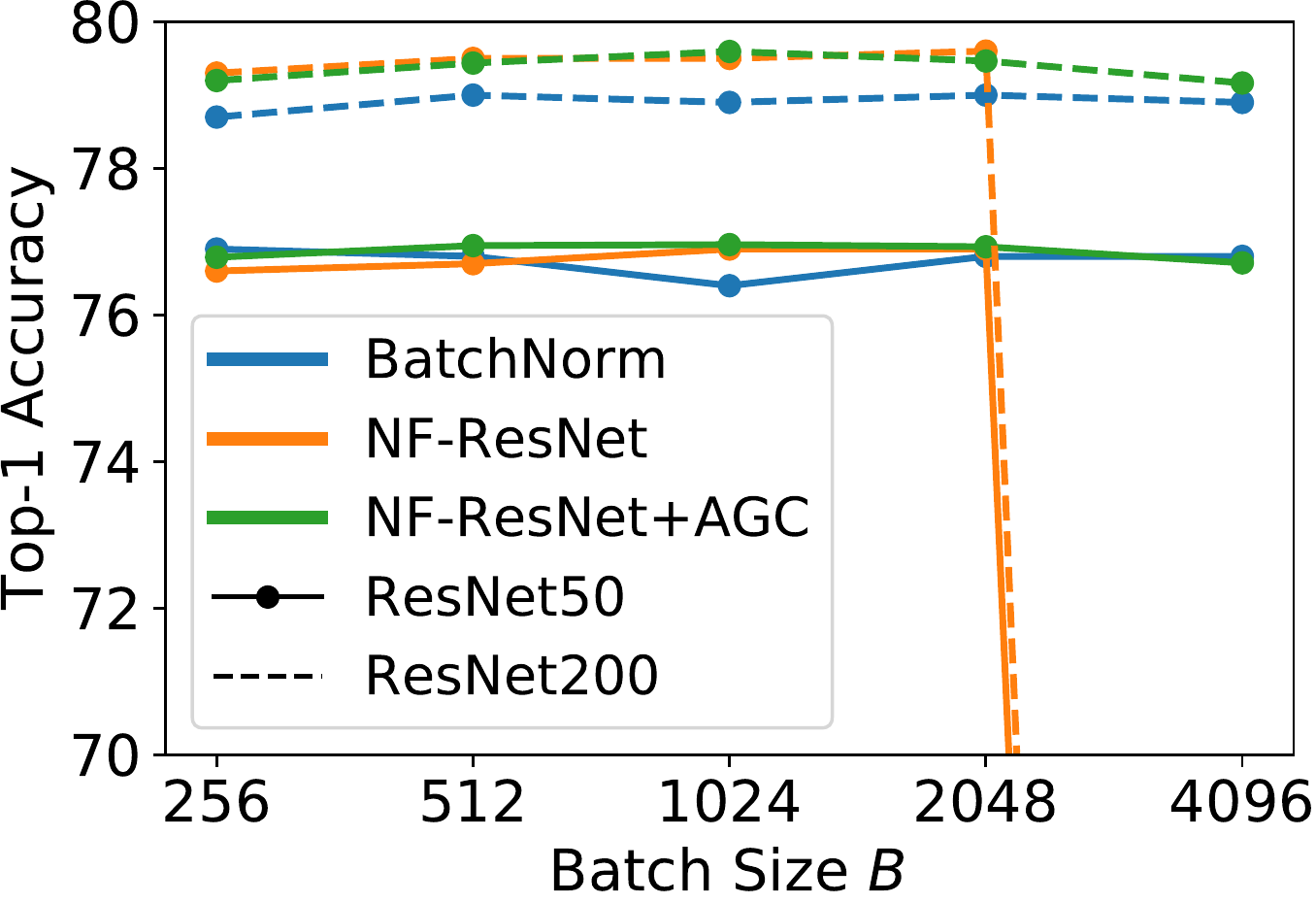}\label{fig:agc_batchsizescaling}}
\subfigure[]{
	\includegraphics[width=0.48\columnwidth]
	{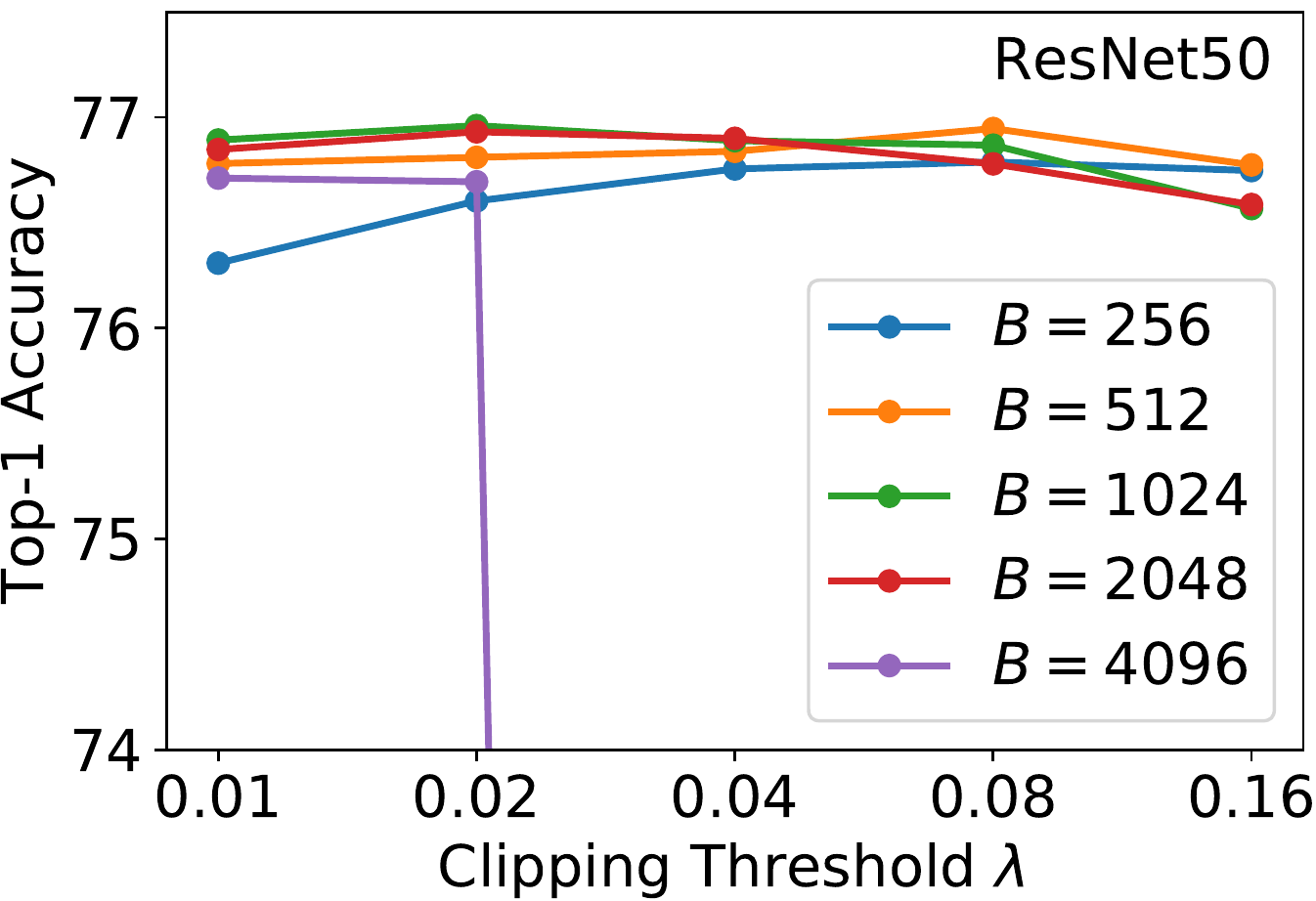}\label{fig:agc_clippingsweep}}
\vskip -4mm
\caption{%
(a) AGC efficiently scales NF-ResNets to larger batch sizes. (b) The performance across different clipping thresholds $\lambda$.} %
\label{fig:agc_batchsize_ablations}
\vskip -2mm
\end{figure}

\subsection{Ablations for Adaptive Gradient Clipping (AGC)}
\label{subsec:agc_ablations}

We now present a range of ablations designed to test the efficacy of AGC. We performed experiments on pre-activation NF-ResNet-50 and NF-ResNet-200 on ImageNet, trained using SGD with Nesterov's Momentum for 90 epochs at a range of batch sizes between 256 and 4096. As in \citet{goyal2017accurate} we use a base learning rate of 0.1 for batch size 256, which is scaled linearly with the batch size. We consider a range of $\lambda$ values [0.01, 0.02, 0.04, 0.08, 0.16]. 

In Figure \ref{fig:agc_batchsizescaling}, we compare batch-normalized ResNets to NF-ResNets with and without AGC. We show test accuracy at the best clipping threshold $\lambda$ for each batch size. We find that AGC helps scale NF-ResNets to large batch sizes while maintaining performance comparable or better than batch-normalized networks on both ResNet50 and ResNet200. 
As anticipated, the benefits of using AGC are smaller when the batch size is small. 
In Figure \ref{fig:agc_clippingsweep}, we show performance for different clipping thresholds $\lambda$ across a range of batch sizes on ResNet50. We see that smaller (stronger) clipping thresholds are necessary for stability at higher batch sizes.
We provide additional ablation details in Appendix \ref{appendix:agc_ablations}.

Next, we study whether or not AGC is beneficial for all layers. Using batch size 4096 and a clipping threshold $\lambda = 0.01$, we remove AGC from different combinations of the first convolution, the final linear layer, and every block in any given set of the residual stages. For example, one experiment may remove clipping in the linear layer and all the blocks in the second and fourth stages. Two key trends emerge: first, it is always better to not clip the final linear layer. Second, it is often possible to train stably without clipping the initial convolution, but the weights of all four stages must be clipped to achieve stability when training at batch size 4096 with the default learning rate of 1.6. For the rest of this paper (and for our ablations in Figure \ref{fig:agc_batchsize_ablations}), we apply AGC to every layer except for the final linear layer.

%% file: Section5_ModelDesign.tex
In the previous section we introduced AGC, a gradient clipping method which allows us to train efficiently with large batch sizes and strong data augmentations. Equipped with this technique, we now seek to design Normalizer-Free architectures with state-of-the-art accuracy and training speed.

The current state of the art on image classification is generally held by the EfficientNet family of models \citep{tan2019efficientnet}, which are based on a variant of inverted bottleneck blocks \citep{sandler2018mobilenetv2} with a backbone and model scaling strategy derived from neural architecture search. These models are optimized to maximize test accuracy while minimizing parameter and FLOP counts, but their low theoretical compute complexity does not translate into improved training speed on modern accelerators. Despite having 10x fewer FLOPS than a ResNet-50, an EffNet-B0 has similar training latency and final performance when trained on GPU or TPU. 

\begin{figure}[t]
	\includegraphics[width=0.99\columnwidth]
	{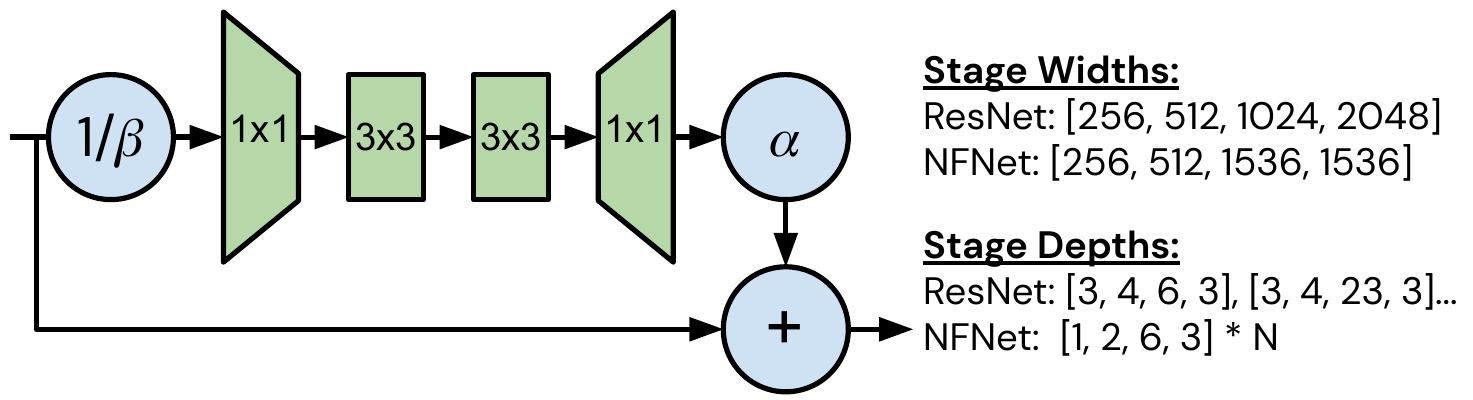}
	\vskip -2mm
	\caption{\label{fig:NFNet_block_small} Summary of NFNet bottleneck block design and architectural differences. See Figure~\ref{fig:NFNet_block_large} in Appendix \ref{appendix:model_details} for more details.} 
	\vskip -4mm
\end{figure}

\begin{table}[t]
\vskip -2mm
\caption{\label{table:nfnet_arch_table} NFNet family depths, drop rates, and input resolutions.}
\input{tables/nfnet_arch_table.tex}
\vskip -3mm
\end{table}

The choice of which metric to optimize-- theoretical FLOPS, inference latency on a target device, or training latency on an accelerator--is a matter of preference, and the nature of each metric will yield different design requirements. In this work we choose to focus on manually designing models which are optimized for training latency on existing accelerators, as in \citet{radosavovic2020designing}. It is possible that future accelerators may be able to take full advantage of the potential training speed that largely goes unrealized with models like EfficientNets, so we believe this direction should not be ignored \citep{hooker2020hardware}, however we anticipate that developing models with improved training speed on current hardware will be beneficial for accelerating research. We note that accelerators like GPU and TPU tend to favor dense computation, and while there are differences between these two platforms, they have enough in common that models designed for one device are likely to train fast on the other.

We therefore explore the space of model design by manually searching for design trends which yield improvements to the pareto front of holdout top-1 on ImageNet against actual training latency on device. This section describes the changes which we found to work well to this end (with more details in Appendix~\ref{appendix:model_details}), while the ideas which we found to work poorly are described in Appendix~\ref{appendix:negative_results}. A summary of these modifications is presented in Figure~\ref{fig:NFNet_block_small}, and the effect they have on holdout accuracy is presented in Table~\ref{table:model_ablation_table0}. %

We begin with an SE-ResNeXt-D model \citep{xie2017aggregated, hu2018squeeze,he2019bag} with GELU activations \citep{hendrycks2016gaussian}, which we found to be a surprisingly strong baseline for Normalizer-Free Networks. %
We make the following changes. 
First, we set the group width (the number of channels each output unit is connected to) in the $3\times3$ convs to 128, regardless of block width. Smaller group widths reduce theoretical FLOPS, but the reduction in compute density means that on many modern accelerators no actual speedup is realized.
On TPUv3 for example, an SE-ResNeXt-50 with a group width of 8 trains at the same speed as an SE-ResNeXt-50 with a group width of 128 unless the per-device batch size is 128 or larger \citep{tpu_performance}, which is often not realizable due to memory constraints. %

Next, we make two changes to the model backbone. First, we note that the default depth scaling pattern for ResNets (e.g., the method by which one increases depth to construct a ResNet101 or ResNet200 from a ResNet50) involves non-uniformly increasing the number of layers in the second and third stages, while maintaining 3 blocks in the first and fourth stages, where `stage' refers to a sequence of residual blocks whose activations are the same width and have the same resolution. We find that this strategy is suboptimal. %
Layers in early stages operate at higher resolution, require more memory and compute, and tend to learn localized, task-general features \citep{krizhevsky2012imagenet}, while layers in later stages operate at lower resolutions, contain most of the model's parameters, and learn more task-specific features \citep{raghu2017svcca}. However, being overly parsimonious with early stages (such as through aggressive downsampling) can hurt performance, since the model needs enough capacity to extract good local features \citep{raghu2017expressive}. It is also desirable to have a simple scaling rule for constructing deeper variants \citep{tan2019efficientnet}. With these principles in mind, we explored several choices of backbone for our smallest model variant, named F0, before settling on the simple pattern $[1, 2, 6, 3]$ (indicating how many bottleneck blocks to allocate to each stage). We construct deeper variants by multiplying the depth of each stage by a scalar $N$, so that, for example, variant F1 has a depth pattern $[2, 4, 12, 6]$, and variant F4 has a depth pattern $[5, 10, 30, 15]$.

\begin{table}[t]
\vskip -2mm
\caption{\label{table:model_ablation_table0}The effect of architectural modifications and data augmentation on ImageNet Top-1 accuracy (averaged over 3 seeds).}
\input{tables/model_ablation_table0.tex}
\vskip -3mm
\end{table}

In addition, we reconsider the default width pattern in ResNets, where the first stage has 256 channels which are doubled at each subsequent stage, resulting in a pattern $[256, 512, 1024, 2048]$. Employing our depth patterns described above, we considered a range of alternative patterns (taking inspiration from \citet{radosavovic2020designing}) but found that only one choice was better than this default: $[256, 512, 1536, 1536]$. This width pattern is designed to increase capacity in the third stage while slightly reducing capacity in the fourth stage, roughly preserving training speed. Consistent with our chosen depth pattern and the default design of ResNets, we find that the third stage tends to be the best place to add capacity, which we hypothesize is due to this stage being deep enough to have a large receptive field and access to deeper levels of the feature hierarchy, while having a slightly higher resolution than the final stage.

We also consider the structure of the bottleneck residual block itself. We considered a variety of pre-existing and novel modifications (see Appendix~\ref{appendix:negative_results}) but found that the best improvement came from adding an additional $3\times3$ grouped conv after the first (with accompanying nonlinearity). This additional convolution minimally impacts FLOPS and has almost no impact on training time on our target accelerators. %

Finally, we establish a scaling strategy to produce model variants at different compute budgets. The EfficientNet scaling strategy \citep{tan2019efficientnet} is to jointly scale model width, depth, and input resolution, which works extremely well for base models with very slim MobileNet-like backbones. %
However we find that width scaling is ineffective for ResNet backbones, consistent with \citet{bello2021lambdanetworks}, who attain strong performance when only scaling depth and input resolution. We therefore also adopt the latter strategy, using the fixed width pattern mentioned above, scaling depth as described above, and scaling training resolution such that each variant is approximately half as fast to train as its predecessor. Following \citet{touvron2019fixing}, we evaluate images at inference at a slightly higher resolution than we train at, chosen for each variant as approximately 33\% larger than the train resolution. We do not fine-tune at this higher resolution.

We also find that it is helpful to increase the regularization strength as the model capacity rises. However  modifying the weight decay or stochastic depth rate was not effective, and instead we scale the drop rate of Dropout \citep{srivastava2014dropout}, following \citet{tan2019efficientnet}. This step is particularly important as our models lack the implicit regularization of batch normalization, and without explicit regularization tend to dramatically overfit. Our resulting models are highly performant and, despite being optimized for training latency, remain competitive with larger EfficientNet variants in terms of FLOPs vs accuracy (although not in terms of parameters vs accuracy), as shown in  Figure~\ref{fig:flops_pareto_front} in Appendix \ref{appendix:experiment_details}.

\subsection{Summary}
\label{subsec:recipe_summary}
Our training recipe can be summarized as follows:
First, apply the Normalizer-Free setup of \citet{brock2021characterizing} to an SE-ResNeXt-D, with modified width and depth patterns, and a second spatial convolution. Second, apply AGC to every parameter except for the linear weight of the classifier layer. For batch size 1024 to 4096, set $\lambda=0.01$,
and make use of strong regularization and data augmentation. See Table~\ref{table:nfnet_arch_table} for additional information on each model variant.

%% file: tables/nfnet_arch_table.tex
\begin{center}\begin{tabular}{l | c | c | c | c}
\toprule [0.15em]
Variant & Depth & Dropout & Train & Test \\
\midrule [0.05em]
F0 & [1, 2, 6, 3] & 0.2 & 192px & 256px \\
F1 & [2, 4, 12, 6] & 0.3 & 224px & 320px \\
F2 & [3, 6, 18, 9] & 0.4 & 256px & 352px \\
F3 & [4, 8, 24, 12] & 0.4 & 320px & 416px \\
F4 & [5, 10, 30, 15] & 0.5 & 384px & 512px \\
F5 & [6, 12, 36, 18] & 0.5 & 416px & 544px \\
F6 & [7, 14, 42, 21] & 0.5 & 448px & 576px \\
\bottomrule[0.15em]
\end{tabular}
\end{center}

%% file: tables/model_ablation_table0.tex
\begin{center}\begin{tabular}{l| c | c | c | c }
\toprule [0.15em]
 & F0 & F1 & F2 & F3 \\
\midrule [0.1em]
Baseline & 80.4 & 81.7 & 82.0 & 82.3 \\
+ Modified Width & 80.9 & 81.8 & 82.0 & 82.3 \\
+ Second Conv & 81.3 & 82.2 & 82.4 & 82.7 \\
+ MixUp & 82.2 & 82.9 & 83.1 & 83.5 \\
+ RandAugment & 83.2 & 84.6 & 84.8 & 85.0 \\
+ CutMix & \textbf{83.6} & \textbf{84.7} & \textbf{85.1} & \textbf{85.7} \\
Default Width + Augs & 83.1 & 84.5 & 85.0 & 85.5 \\
\bottomrule[0.15em]
\end{tabular}
\end{center}

%% file: Section6_Experiments.tex
\begin{table*}[t]  %
\input{tables/imagenet_results_table.tex}

\caption{\label{table:imagenet_results_table} ImageNet Accuracy comparison for NFNets and a representative set of models, including SENet \citep{hu2018squeeze}, LambdaNet, \citep{bello2021lambdanetworks}, BoTNet \citep{srinivas2021bottleneck}, and DeIT \cite{touvron2020training}. Except for results using SAM, our results are averaged over three random seeds. Latencies are given as the time in milliseconds required to perform a single full training step on TPU or GPU (V100).}
\end{table*}

\subsection{Evaluating NFNets on ImageNet}

We now turn our attention to evaluating our NFNet models on ImageNet, beginning with an ablation of our architectural modifications when training for 360 epochs at batch size 4096. We use Nesterov's Momentum with a momentum coefficient of 0.9, AGC as described in Section~\ref{sec:UAGC} with a clipping threshold of 0.01, and a learning rate which linearly increases from 0 to 1.6 over 5 epochs, before decaying to zero with cosine annealing \citep{loshchilov2017decoupled}. From the first three rows of Table~\ref{table:model_ablation_table0}, we can see that the two changes we make to the model each result in slight improvements to performance with only minor changes in training latency (See Table~\ref{table:model_ablation_table_detailed} in the Appendix for latencies).

Next, we evaluate the effects of progressively adding stronger augmentations, combining MixUp \citep{zhang2017mixup}, RandAugment (RA, \cite{cubuk2020randaugment}) and CutMix \citep{yun2019cutmix}. We apply RA with 4 layers and scale the magnitude with the resolution of the images, following \citet{cubuk2020randaugment}. We find that this scaling is particularly important, as if the magnitude is set too high relative to the image size (for example, using a magnitude of 20 on images of resolution 224) then most of the augmented images will be completely blank. See Appendix~\ref{appendix:experiment_details} for a complete description of these magnitudes and how they are selected. We show in Table \ref{table:model_ablation_table0} that these data augmentations substantially improve performance. Finally, in the last row of Table~\ref{table:model_ablation_table0}, we additionally present the performance of our full model ablated to use the default ResNet stage widths, demonstrating that our slightly modified pattern in the third and fourth stages does yield improvements under direct comparison. 

For completeness, in Table~\ref{table:model_ablation_table_detailed} of the Appendix we also report the performance of our model architectures when trained with batch normalization instead of the NF strategy. These models achieve slightly lower test accuracies than their NF counterparts and they are between 20\% and 40\% slower to train, even when using highly optimized batch normalization implementations without cross-replica syncing. Furthermore, we found that the larger model variants F4 and F5 were not stable when training with batch normalization, with or without AGC. We attribute this to the necessity of using bfloat16 training to fit these larger models in memory, which may introduce numerical imprecision that interacts poorly with the computation of batch normalization statistics.

We provide a detailed summary of the size, training latency (on TPUv3 and V100 with tensorcores), and ImageNet validation accuracy of six model variants, NFNet-F0 through F5, along with comparisons to other models with similar training latencies, in Table~\ref{table:imagenet_results_table}. Our NFNet-F5 model attains a top-1 validation accuracy of 86.0\%, improving over the previous state of the art, 
EfficientNet-B8 with MaxUp \citep{gong2020maxup}
by a small margin, and our NFNet-F1 model matches the 84.7\% of EfficientNet-B7 with RA \citep{cubuk2020randaugment}, while being 8.7 times faster to train. See Appendix~\ref{appendix:experiment_details} for details of how we measure training latency.

Our models also benefit from the recently proposed Sharpness-Aware Minimization (SAM, \citep{foret2021sharpnessaware}). SAM is not part of our standard training pipeline, as by default it doubles the training time and typically can only be used for distributed training. However we make a small modification to the SAM procedure to reduce this cost to 20-40\% increased training time (explained in Appendix~\ref{appendix:experiment_details}) and employ it to train our two largest model variants, resulting in an NFNet-F5 that attains 86.3\% top-1, and an NFNet-F6 that attains 86.5\% top-1, substantially improving over the existing state of the art on ImageNet without extra data.

Finally, we also evaluated the performance of our data augmentation strategy on EfficientNets. We find that while RA strongly improves EfficientNets' performance over baseline augmentation, increasing the number of layers beyond 2 or adding MixUp and CutMix does not further improve their performance, suggesting that our performance improvements are difficult to obtain by simply using stronger data augmentations. We also find that using SGD with cosine annealing instead of RMSProp \citep{tieleman2012rmsprop} with step decay severely degrades EfficientNet performance, indicating that our performance improvements are also not simply due to the selection of a different optimizer.

\subsection{Evaluating NFNets under Transfer}

\begin{table}[t]
\vskip -2mm
\caption{ImageNet Transfer Top-1 accuracy after  pre-training.}
\label{table:jft_resnets_table}
\input{tables/jft_resnets_table.tex}
\vskip -3mm
\end{table}

Unnormalized networks do not share the implicit regularization effect of batch normalization, and on datasets like ImageNet \citep{ILSVRC2015} they tend to overfit unless explicitly regularized \citep{zhang2019fixup, de2020batch, brock2021characterizing}. However when pre-training on extremely large scale datasets, such regularization may not only be unnecessary, but also harmful to performance, reducing the model's ability to devote its full capacity to the training set. We hypothesize that this may make Normalizer-Free networks naturally better suited to transfer learning after large-scale pre-training, and investigate this via pre-training on a large dataset of 300 million labeled images.

We pre-train a range of batch normalized and NF-ResNets for 10 epochs on this large dataset, then fine-tune all layers on ImageNet simultaneously, using a batch size of 2048 and a small learning rate of 0.1 with cosine annealing for 15,000 steps, for input image resolutions in the range [224, 320, 384]. As shown in Table~\ref{table:jft_resnets_table}, Normalizer-Free networks outperform their Batch-Normalized counterparts in every single case, typically by a margin of around 1\% absolute top-1. This suggests that in the transfer learning regime, removing batch normalization can directly benefit final performance.

We perform this same experiment using our NFNet models, pre-training an NFNet-F4 and a slightly wider variant which we denote NFNet-F4+ (see Appendix~\ref{appendix:model_details}).
As shown in Table~\ref{table:jft_full_results_table} of the appendix, with 20 epochs of pre-training our NFNet-F4+ attains an ImageNet top-1 accuracy of 89.2\%. This is the second highest validation accuracy achieved to date with extra training data, second only to a strong recent semi-supervised learning baseline \citep{pham2020meta}, and the highest accuracy achieved using transfer learning.

%% file: tables/imagenet_results_table.tex
\begin{center}\begin{tabular}{l| c c | c c | c c}
\toprule [0.15em]
Model & \#FLOPs & \#Params & Top-1 & Top-5 & TPUv3 Train & GPU Train \\
\midrule [0.1em]
ResNet-50 & $4.10$B & $26.0$M & $78.6$ & $94.3$ & $41.6$ms & $35.3$ms \\
EffNet-B0 & $0.39$B & $5.3$M & $77.1$ & $93.3$ & $51.1$ms & $44.8$ms \\
SENet-50 & $4.09$B & $28.0$M & $79.4$ & $94.6$ & $64.3$ms & $59.4$ms \\
\textbf{NFNet-F0} & $\textbf{12.38B}$ & $\textbf{71.5M}$ & $\textbf{83.6}$ & $\textbf{96.8}$ & $\textbf{73.3ms}$ & $\textbf{56.7ms}$ \\
\midrule
EffNet-B3 & $1.80$B & $12.0$M & $81.6$ & $95.7$ & $129.5$ms & $116.6$ms \\
LambdaNet-152 & $-$ & $51.5$M & $83.0$ & $96.3$ & $138.3$ms & $135.2$ms \\
SENet-152 & $19.04$B & $66.6$M & $83.1$ & $96.4$ & $149.9$ms & $151.2$ms \\
BoTNet-110 & $10.90$B & $54.7$M & $82.8$ & $96.3$ & $181.3$ms & $-$ \\
\textbf{NFNet-F1} & $\textbf{35.54B}$ & $\textbf{132.6M}$ & $\textbf{84.7}$ & $\textbf{97.1}$ & $\textbf{158.5ms}$ & $\textbf{133.9ms}$ \\
\midrule
EffNet-B4 & $4.20$B & $19.0$M & $82.9$ & $96.4$ & $245.9$ms & $221.6$ms \\
BoTNet-128-T5 & $19.30$B & $75.1$M & $83.5$ & $96.5$ & $355.2$ms & $-$ \\
\textbf{NFNet-F2} & $\textbf{62.59B}$ & $\textbf{193.8M}$ & $\textbf{85.1}$ & $\textbf{97.3}$ & $\textbf{295.8ms}$ & $\textbf{226.3ms}$ \\
\midrule
SENet-350 & $52.90$B & $115.2$M & $83.8$ & $96.6$ & $593.6$ms & $-$ \\
EffNet-B5 & $9.90$B & $30.0$M & $83.7$ & $96.7$ & $450.5$ms & $458.9$ms \\
LambdaNet-350 & $-$ & $105.8$M & $84.5$ & $97.0$ & $471.4$ms & $-$ \\
BoTNet-77-T6 & $23.30$B & $53.9$M & $84.0$ & $96.7$ & $578.1$ms & $-$ \\
\textbf{NFNet-F3} & $\textbf{114.76B}$ & $\textbf{254.9M}$ & $\textbf{85.7}$ & $\textbf{97.5}$ & $\textbf{532.2ms}$ & $\textbf{524.5ms}$ \\
\midrule
LambdaNet-420 & $-$ & $124.8$M & $84.8$ & $97.0$ & $593.9$ms & $-$ \\
EffNet-B6 & $19.00$B & $43.0$M & $84.0$ & $96.8$ & $775.7$ms & $868.2$ms \\
BoTNet-128-T7 & $45.80$B & $75.1$M & $84.7$ & $97.0$ & $804.5$ms & $-$ \\
\textbf{NFNet-F4} & $\textbf{215.24B}$ & $\textbf{316.1M}$ & $\textbf{85.9}$ & $\textbf{97.6}$ & $\textbf{1033.3ms}$ & $\textbf{1190.6ms}$ \\
\midrule
EffNet-B7 & $37.00$B & $66.0$M & $84.7$ & $97.0$ & $1397.0$ms & $1753.3$ms \\
DeIT 1000 epochs & $-$ & $87.0$M & $85.2$ & $-$ & $-$ & $-$ \\
EffNet-B8+MaxUp & $62.50$B & $87.4$M & $85.8$ & $-$ & $-$ & $-$ \\
\textbf{NFNet-F5} & $\textbf{289.76B}$ & $\textbf{377.2M}$ & $\textbf{86.0}$ & $\textbf{97.6}$ & $\textbf{1398.5ms}$ & $\textbf{2177.1ms}$ \\
\midrule
NFNet-F5+SAM & $289.76$B & $377.2$M & $86.3$ & $97.9$ & $1958.0$ms & $-$ \\
\textbf{NFNet-F6+SAM} & $\textbf{377.28B}$ & $\textbf{438.4M}$ & $\textbf{86.5}$ & $\textbf{97.9}$ & $\textbf{2774.1ms}$ & $-$ \\
\bottomrule[0.15em]
\end{tabular}
\end{center}

%% file: tables/jft_resnets_table.tex
\begin{center}\begin{tabular}{l| c | c | c}
\toprule [0.15em]
 & 224px & 320px & 384px \\
\midrule [0.1em]
BN-ResNet-50 & 78.1 & 79.6 & 79.9 \\
NF-ResNet-50 & \textbf{79.5} & \textbf{80.9} & \textbf{81.1} \\
\midrule [0.05em]
BN-ResNet-101 & 80.8 & 82.2 & 82.5 \\
NF-ResNet-101 & \textbf{81.4} & \textbf{82.7} & \textbf{83.2} \\
\midrule [0.05em]
BN-ResNet-152 & 81.8 & 83.1 & 83.4 \\
NF-ResNet-152 & \textbf{82.7} & \textbf{83.6} & \textbf{84.0} \\
\midrule [0.05em]
BN-ResNet-200 & 81.8 & 83.1 & 83.5 \\
NF-ResNet-200 & \textbf{82.9} & \textbf{84.1} & \textbf{84.3} \\
\bottomrule[0.15em]
\end{tabular}
\end{center}

%% file: Section7_Conclusion.tex
We show for the first time that image recognition models, trained without normalization layers, can not only match the classification accuracies of the best batch normalized models on large-scale datasets but also substantially exceed them, while still being faster to train. To achieve this, we introduce Adaptive Gradient Clipping, a simple clipping algorithm which stabilizes large-batch training and enables us to optimize unnormalized networks with strong data augmentations. Leveraging this technique and simple architecture design principles, we develop a family of models which attain state-of-the-art performance on ImageNet without extra data, while being substantially faster to train than competing approaches. We also show that Normalizer-Free models are better suited to fine-tuning after pre-training on very large scale datasets than their batch-normalized counterparts.

%% file: Appendix_Experiment_Details.tex
\subsection{ImageNet Experiment Settings}
\label{subsec:appendix_imagenet_details}

\begin{figure}[ht]
	\includegraphics[width=0.99\columnwidth]{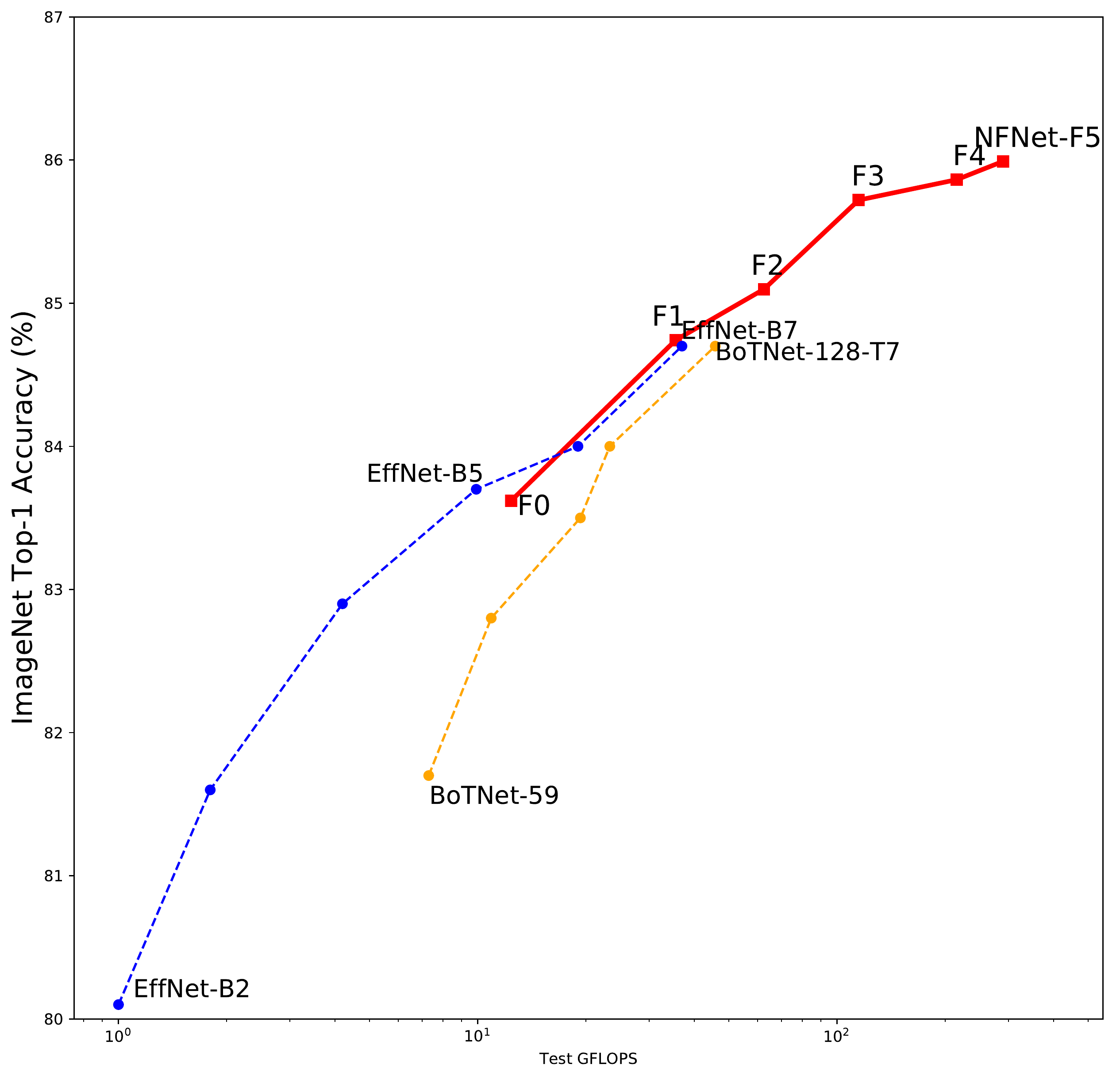}

\caption{\label{fig:flops_pareto_front}\textbf{ImageNet Validation Accuracy vs. Test GFLOPs.} All numbers are single-model, single crop. Our NFNet models are competitive with large EfficientNet variants for a given FLOPs budget, despite being optimized for training latency.}
\end{figure}

\begin{table*}[t]

\input{tables/jft_big_table.tex}
\caption{\label{table:jft_full_results_table} Comparing ImageNet transfer performance for models which use extra data for large-scale pre-training. Meta-Psuedo-Labels results are from \citet{pham2020meta}, ViT results are from \citet{dosovitskiy2021an}, BiT results are from \citet{kolesnikov2019large}. Noisy Student results \citep{xie2020self} are taken from the improved versions reported in \citet{foret2021sharpnessaware} which employ SAM. IG-940M  \citep{mahajan2018exploring} results are taken from the improved versions reported in \citet{touvron2019fixing}.}
\end{table*}

For ImageNet experiments \citep{ILSVRC2015}, we train on the standard ILSVRC2012 training split, which comprises 1281167 images from 1000 classes. Our baseline training preprocessing follows \citet{szegedy2016rethinking}, with distorted bounding box crops and random horizontal flips \citep{simonyan2015vgg}, with all other augmentations being applied in addition to this. We train using the categorical softmax cross-entropy loss with label smoothing of 0.1 \citep{szegedy2016rethinking}, and optimize our networks using stochastic gradient descent \citep{robbins1951A} with Nesterov's momentum \citep{nesterov1983, sutskever2013importance}, using a momentum coefficient of 0.9.
Our training code is available at \url{https://github.com/deepmind/deepmind-research/tree/master/nfnets}, and is written using numpy \citep{harris2020numpy}, JAX \citep{jax2018github}, Haiku \citep{haiku2020github}, and the DeepMind JAX Ecosystem \citep{deepmind2020jax}.

We employ weight decay in the standard style (not decoupled as in \citet{loshchilov2017decoupled}), with a weight decay coefficient of $2\times10^{-5}$ for NFNets. Critically, weight decay is not applied to the affine gains or biases in the weight-standardized convolutional layers, or to the SkipInit gains. We apply a Dropout rate specific to each NFNet variant as in \citet{tan2019efficientnet}, and use Stochastic Depth with a rate of 0.25 for all variants, again similar to \citet{tan2019efficientnet}.

We use a learning rate which warms up from 0 to its maximal value over the first 5 epochs, where the maximal value is chosen as $0.1\times B/256$, with $B$ the batch size, following \citet{goyal2017accurate}. After warmup, the learning rate is annealed to zero with cosine decay over the rest of training \citep{loshchilov2016sgdr}. We employ AGC with $\lambda=0.01$ and $\epsilon=10^{-3}$ for every parameter except the fully-connected weight of the linear classifier layer.

By default, we train with a batch size of 4096 for 360 epochs, a common training schedule which has the same number of total training steps (roughly 112,000) as training with a batch size of 1024 for 90 epochs. We found that training for longer sometimes improved results, but that this was not always consistent across models or training settings; all results reported in this work employ the 360 epoch schedule. Unlike \citet{tan2019efficientnet} we do not perform early stopping.

We employ an exponential moving average of the model parameters (similar to Polyak averaging \citep{polyak1964some}), with a decay rate of 0.99999 which, following \citet{tan2019efficientnet}, follows a warmup schedule where the decay is equal to $ min(0.99999, \frac{1 + t}{10 + t})$. %

We train on TPU using bfloat16 activations to save memory and improve speed. This means that we keep the parameters and optimizer state (the momentum buffer) in float32, but compute activations and gradients in bfloat16 during forward- and backpropagation. We cast the logits to float32 before computing the loss to aid numerical stability. We cast gradients back to float32 before summing them across devices, which helps prevent compounding accumulation error and ensures the parameter update is computed in float32.

For evaluation we follow the most common style of single-crop preprocessing: we resize the raw image (with bicubic interpolation) to be 32 pixels larger than the target resolution, then crop to the target resolution \citep{simonyan2015vgg}. While this is the most commonly employed variant, we note that an alternative method exists where a padded center crop is taken and then resized to the target resolution \citep{szegedy2016inception, tan2019efficientnet}. We find this alternative to work marginally worse than the standard choice of resizing before cropping. No test time augmentation, multi-crop evaluation, or model ensembling is applied.

\subsection{Measuring Training Latency}
\label{subsec:appendix_measuring_speed}

We measure training latency as the actual observed wallclock time required to perform a training step at a given per-device batch size. To accomplish this, we run the full training loop for 5000 steps, then take the median time required to perform a single training step. We choose the median as the mean would also incorporate the initial speed ramp-up at the beginning of training, so the median is more robust to these types of variations during measurement and better reflects the speed observed during a full training run. We remove dataloading as a consideration by having the training loop operate on tensors which are already loaded onto the device. This is consistent with how we train NFNets in practice, since our data pipeline is optimized to ensure we are never input-bound.

For measuring speed on TPUv3, we run on 32 devices with a batch size of 32 per device, and sync gradients between replicas, meaning that our training latency is representative of the actual speed we can obtain in practice with distributed training. We employ bfloat16 training for all models, as described above. For some of our larger models, this batch size of 32 per device does not fit into the 16GB of device memory, so we allow the compiler to engage automatic rematerialization (also known as gradient checkpointing). Additional speed may be obtainable by careful tuning of manual rematerialization.

For measuring speed on GPU, we run on a single V100 GPU using float16 training to engage the card's tensorcores, which strongly accelerates training. Unlike TPUv3, we do not consider the cost of cross-device communication for GPU, which will vary substantially depending on the hardware configuration of the interlinks available to the user. As with TPUv3, some of our models do not fit in memory at this batch size, but we instead employ gradient accumulation to mimic the full batch size. This appears to be less efficient than rematerialization for large models (specifically for our F5 variant and for EfficientNet-B7), so we expect that manually applying rematerialization would potentially yield GPU speedups in this case, but require extra engineering effort.

We report results from our own measurements for all models except for SENets \citep{hu2018squeeze}, BoTNets \citep{srinivas2021bottleneck}, and DeIT \citep{touvron2020training}, which we instead borrow from \citet{srinivas2021bottleneck}. We report slightly different training latencies for small EfficientNet variants because we report the wallclock time, whereas \citet{srinivas2021bottleneck} report the ``compute time'' which will ignore cross-device communication. For very small models the inter-device communication costs can be non-negligible relative to the compute time, especially for EfficientNets which employ cross-replica batch normalization.
For larger models this cost is generally negligible on hardware like TPUv3 with very fast interconnects, so in practice one can expect that the compute time for models like BoTNets will be the same regardless of the reporting methodology used.

\subsection{Augmentations}
\label{subsec:appendix_aug_details}

Our full NFNet training recipe applies ``baseline'' preprocessing (sampling distorted bounding boxes and applying random horizontal flips), RandAugment (RA, \citet{cubuk2020randaugment}), which we apply to all images in a batch, MixUp \citep{zhang2017mixup}, which we apply to half the images in a batch with $\alpha=0.2$, and CutMix \citep{yun2019cutmix}, which we apply to the other half of the images in the batch. 

Following \citet{qin2020resizemix} we apply RandAugment after applying MixUp or CutMix. We apply RA with 4 layers (meaning 4 augmentations are chosen), which is substantially stronger than the common default of 2 layers, and following \citet{cubuk2020randaugment} we pick the magnitude of the RA augmentation based on the training resolution of the images. If the augmentation magnitude is set too high relative to the image resolution, then certain operations (such as shearing) can result in many images being completely blank, which will impede training. For NFNet variants F0 through F6, the chosen RA magnitudes are $[5, 10, 10, 15, 15, 15, 15]$, respectively.

The combination of MixUp, CutMix, and RA results in an intense level of augmentation which progressively benefits NFNets, but does not appear to benefit other models like EfficientNets over a baseline of just using well-tuned RA. We hypothesize that this is because our models lack the implicit regularization of batch normalization, and similar to how they are more amenable to large-scale pre-training, they are accordingly also more amenable to stronger data augmentations.

\subsection{Accelerating Sharpness-Aware Minimization}
\label{subsec:appendix_fast_SAM}

Sharpness-Aware Minimization (SAM, \citet{foret2021sharpnessaware}) has been shown to improve the performance of various classifier models by seeking flat minima which are hypothesized to generalize better. However, by default it is expensive to apply as it requires two evaluations of the gradient: one for a step of gradient ascent to attain ``noised'' parameters, and then one to attain the gradients with respect to the noised parameters, which are used to update the actual parameters. We experimented with ameliorating this cost by only employing 20\% of the batch to compute the gradients for the ascent step, which we found to result in equivalent performance while only increasing the training latency by 20\%-40\% instead of by 100\%. We also tried using SAM where the batch of data used to compute the ascent step was a different batch from the one used to compute the descent step, but found that this destroyed all the benefits of SAM. This indicates that it is necessary for the ascent step to be computed using the same batch (or a subset thereof) as is used to compute the descent step. As noted in \citet{foret2021sharpnessaware}, we found that SAM worked best in a distributed setup where the gradients used for the ascent step are \textit{not} synced between replicas (meaning a separate copy of the ``noised'' parameters is kept on each replica and used to compute the local descent gradients). We note that this phenomenon can also be mimicked on fewer devices, or a single device, by employing gradient accumulation (iteratively computing noised parameters and then accumulating the gradients to be used for descent).

\subsection{Large Scale Pre-Training Details}
\label{subsec:appendix_pretraining}

\input{pretraining_public.tex}

%% file: tables/jft_big_table.tex
\centering
\begin{tabular}{l c c c c}
\toprule
Model & \#FLOPS & \#Params & ImageNet Top-1 & TPUv3-core-days \\
\midrule 
NFNet-F4+ (ours)  & $367$B & $527$M & $89.2$ & $1.86$k \\
NFNet-F4 (ours)  & $215$B & $316$M & $89.2$ & $3.7$k \\
EffNet-L2 + Meta Pseudo Labels & -  & $480$M & $\mathbf{90.2}$ & $22.5$k  \\
EffNet-L2 + NoisyStudent + SAM & - & $480$M & 88.6 & $12.3$k  \\
ViT-H/14          & - & $632$M & \valstd{88.55}{0.04} & $2.5$k \\
ViT-L/16          & - & $307$M &  \valstd{87.76}{0.03} & $0.68$k \\
BiT-L ResNet152x4 & - & $928$M & \valstd{87.54}{0.02} & $9.9$k  \\
ResNeXt-101 32x48d (IG-940M) & - & $829$M & $86.4$ & - \\

\bottomrule
\end{tabular}

%% file: pretraining_public.tex
Our large scale pre-training is performed on JFT-300m \citep{sun17revisiting}, a dataset of 300 million labeled images spanning roughly 18,000 classes. We pre-train all models at resolution 224 (regardless of the native model resolution for a given NFNet variant) using the same optimizer settings as for our ImageNet experiments (as described in Appendix~\ref{subsec:appendix_imagenet_details}) with the exception of using a smaller weight decay ($10^{-5}$ for BN and NF-ResNets, and $10^{-6}$ for all NFNet models). We briefly tried pre-training at larger image resolutions and found that this was not worth the added pre-trainining expense. We do not use any augmentations except for baseline random crops and flips, nor do we use any exponential moving averages during pre-training.

For ResNet models, we pre-train with a batch size of 1024 for 10 epochs using a learning rate of 0.4 following \citet{goyal2017accurate}, which is warmed up over 5,000 steps and then decayed to zero with cosine annealing through the rest of training. We fine-tune ResNets on ImageNet with a batch size of 2048 for 15,000 steps using a learning rate of 0.1 (again employing a 5000 step warmup and cosine decay, but not applying the batch size scaling of \citet{goyal2017accurate}), no weight decay, no DropOut, and no Stochastic Depth. For fine-tuning we apply EMA with decay 0.9999 and the decay warmup described above. Due to the expense of this experiment we only run a single random seed for each model (fine-tuning three separate times at each of the fine-tune resolutions of 224, 320, and 384 pixels).

We find, contrary to \citep{dosovitskiy2021an}, that a large weight decay is harmful during pre-training, and that instead very small weight decays are important so that the models are not constrained when trying to capture the information in a large scale dataset. Contrary to \citet{dosovitskiy2021an} we also find that Adam is not as performant as SGD in this setting. We believe this reflects in the fact that our baseline batch-normalized ResNets substantially outperform the baselines reported in \citet{dosovitskiy2021an} despite otherwise similar pre-training and fine-tuning configurations. For reference, \citet{dosovitskiy2021an} report a ResNet-50 transfer accuracy of 77.54\% when fine-tuned at 384px resolution, whereas we obtain an accuracy of 79.9\% in the same setting for BN-ResNet-50 and 81.1\% for NF-ResNet-50. The full set of accuracies for these ResNet models is available in Table~\ref{table:jft_resnets_table}. We recommend future work on large-scale pre-training to begin with a weight decay of zero and consider lightly increasing it, rather than starting with a large value of weight decay and experimenting with decreasing it.

For NFNet models, we pre-train with a batch size of 4096. For NFNet-F4, we pre-train for 40 epochs, and for NFNet-F4+ we pre-train for 20 epochs. The F4+ model is a wider variant, constructed from the F4 model by using a channel pattern of $[384, 768, 2048, 2048]$ instead of $[256, 512, 1536, 1536]$ and keeping all other hyper-parameters the same. We find that both models obtain about the same training latency (around 830ms per step when training with a per-core batch size of 32), but that the F4 model needs the additional pre-training time to reach the same final performance as the F4+ model. This indicates that (given sufficient pre-training data) it is more efficient to train larger models with a shorter epoch budget than to train smaller models for longer, consistent with the observations in \citep{kaplan2020scaling}. 

We fine-tune NFNet models for 15,000 steps at a batch size of 2048 using a learning rate of 0.1, which is warmed up from zero over 5000 steps, then annealed to zero with cosine decay through the rest of training. We use SAM with $\rho=0.05$, weight decay of $10^{-5}$, a DropOut rate of 0.25, and a stochastic depth rate of 0.1. We found that we could obtain similar results using the same regularization setup as for ResNets (no weight decay, DropOut, or Stochastic Depth) but that this mild degree of augmentation was slightly more performant. As with our ResNet fine-tuning we employ an exponential moving average of the parameters with EMA decay warmup. The results of this experiment, compared against other models which are pre-trained on large scale datasets, are available in Table~\ref{table:jft_full_results_table}.

%% file: Appendix_WhyBN_Bad.tex
Batch normalization provides a range of benefits, which we discussed in Section \ref{subsec:batchnorm_benefits} of the main text, but it also has a number of disadvantages that motivated this work on normalizer-free networks. We discussed some of the disadvantages of batch normalization in Section \ref{sec:intro}. In addition, here we enumerate some documented errors and challenges in the implementation of batch normalization in popular frameworks and published work. A number of these errors are identified by \citet{pham2019cradle}, an academic paper on automated testing which discovers two such implementation errors in Keras and one in the CNTK toolkit. 

One example is a long-standing bug in certain versions of Keras, whose consequence is that even if a user sets the batch normalization layers to testing mode (as is common when freezing the layers for fine-tuning for downstream tasks) the batch normalization statistics will continue to update, contrary to user expectations. This implementation error is raised in \href{https://github.com/keras-team/keras/issues/7051}{in this github issue} and \href{https://github.com/keras-team/keras/issues/12400}{this github issue.}

The discrepancy between batch normalization train and test behavior has had direct impact several times in previous work. For examples, both DCGAN \citep{radford2016dcgan} and SAGAN \citep{zhang2019self} reported results and released code where batch normalization was run in training mode at test time \href{https://github.com/brain-research/self-attention-gan/commit/7702dc5b5f7c58c14c860232505a6e18f8fb720d}{as noted here} and \href{https://github.com/Newmu/dcgan_code/blob/ee12b2d15a3856794b8dae77d1eb263c67c36e47/faces/train_uncond_dcgan.py\#L111}{here},\footnote{Note that no `u' or `s' values are passed into the batch normalization op \href{https://github.com/Newmu/dcgan_code/blob/ee12b2d15a3856794b8dae77d1eb263c67c36e47/lib/ops.py\#L52-L57}{here}, meaning that running statistics are not accumulated.} and consequently their reported results depend on the batch size used to generate samples.

Subtle differences in batch normalization implementations can also hamper reproducibility. For example, the EfficientNet training code uses a form of cross-replica BatchNorm where the number of devices used to compute statistics varies nonlinearly with the total number of devices (\href{https://github.com/tensorflow/tpu/blob/6f9c87c1215a67fd97da7272eff247e53f266e80/models/official/efficientnet/utils.py\#L118-L140}{as seen here}), and consequently, even given the same code, exact reproduction can be difficult without access to the same hardware. Additionally, the EfficientNet code takes a moving average of the running batch normalization statistics, which in practice means that it takes a moving average of a moving average, compounding the averaging horizon in a way that may be unexpected.

As discussed in the main text, breaking the independence between training examples causes issues in contrastive learning setups like SimCLR \citep{chen2020simple} and MoCo \citep{he2020momentum}.Both models have to deal with the potential for intra-batch information leakage negatively impacting the contrastive objective. MoCo seeks to resolve this by shuffling examples between devices when computing batch statistics, which introduces implementation complexity and makes it challenging to exactly reproduce their results on different hardware. SimCLR seeks to resolve this via the use of cross-replica batch normalization.

%% file: Appendix_Model_Details.tex
Our NFNet model is a modified SE-ResNeXt-D \citep{he2016resnets, he2016identity, xie2017aggregated, hu2018squeeze, he2019bag}. The input to the model is an $H\times W$ RGB image which has been normalized by the per-channel mean / standard deviation from the entire ImageNet \citep{ILSVRC2015} training set, as is standard in most image classifiers. The model has an initial ``stem'' comprised of a $3\times3$ stride 2 convolution with 16 channels, two $3\times3$ stride 1 convolutions with 32 channels and 64 channels respectively, and a final $3\times3$ stride 2 convolution with 128 channels. A nonlinearity is placed in between each convolution in the stem, but importantly not after the final convolution in the stem. By default we use GELU \citep{hendrycks2016gaussian}, although most common nonlinearities like ReLU or SiLU appear to have similar performance. All our nonlinearities are rescaled to be approximately variance-preserving following \citet{brock2021characterizing} using a fixed scalar gain, for which we provide reference values in 
our source code.

\begin{table*}[t]
\input{tables/model_ablation_table_detailed.tex}
\caption{\label{table:model_ablation_table_detailed} Detailed Model ablation table. Each entry reports ImageNet Top-1 on the left, and TPUv3 training latency on the right.}
\end{table*}

\begin{figure}[t]
	\includegraphics[width=0.99\columnwidth]
	{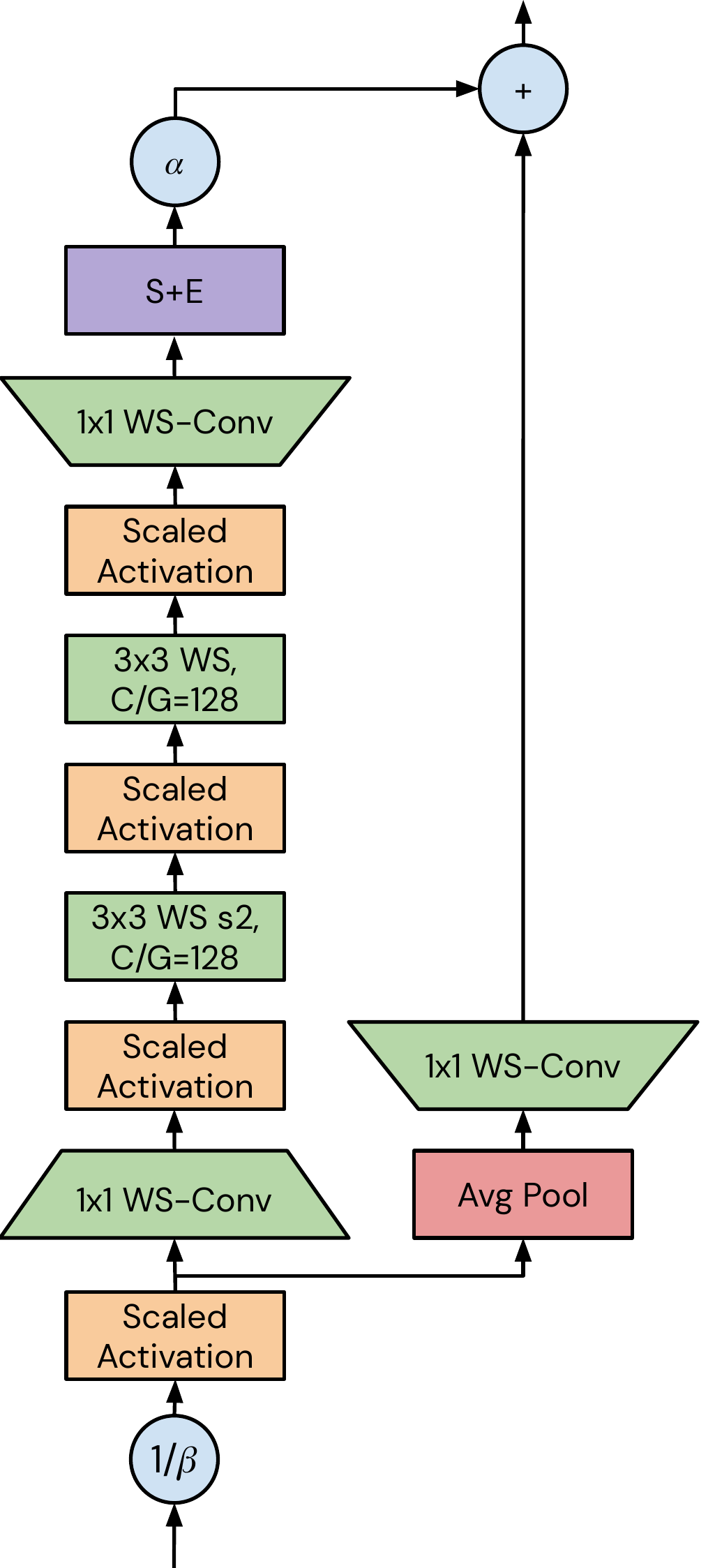}
	\caption{\label{fig:NFNet_block_large} Detailed view of an NFNet transition block. The bottleneck ratio is $0.5$, while the group width (the number of channels per group, $C/G$) in the $3\times3$ convolutions is fixed at 128 regardless of the number of channels. Note that in this block, the skip path takes in the signal after the variance downscaling with $\beta$ and the scaled nonlinearity.} 
\end{figure}

\begin{figure}[t]
	\includegraphics[width=0.99\columnwidth]
	{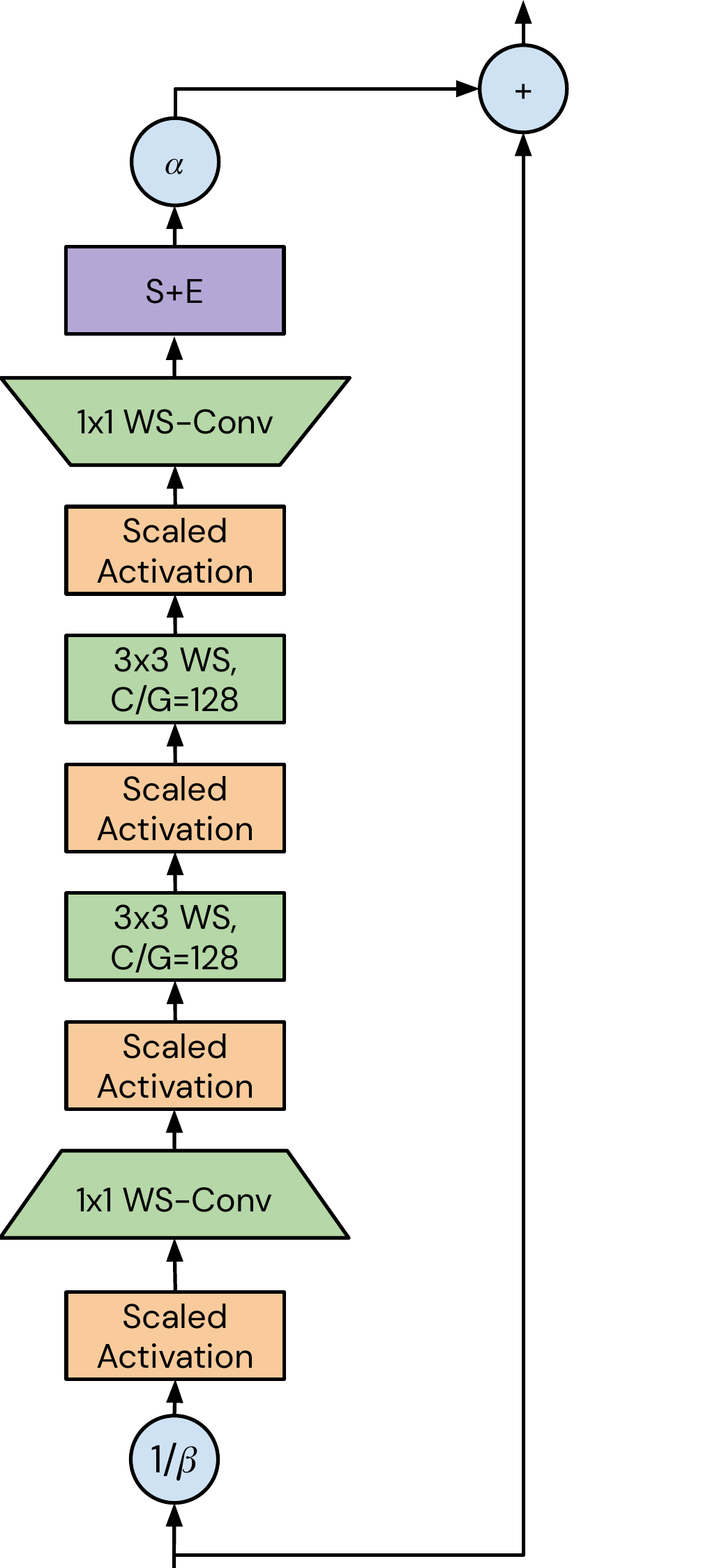}
	\caption{\label{fig:NFNet_block_nontransition} Detailed view of an NFNet non-transition block. The bottleneck ratio is $0.5$, while the group width (the number of channels per group, $C/G$) in the $3\times3$ convolutions is fixed at 128 regardless of the number of channels. Note that in this block, the skip path takes in the signal before the variance downscaling with $\beta$.} 
\end{figure}

Following the stem are four residual ``stages'', where the number of blocks per stage is $[1, 2, 6, 3]$ for our baseline F0 variant, and each subsequent variant has this number multiplied by $N$ (where $N=1$ for F0). The residual stages begin with a ``transition'' block (as shown in Figure~\ref{fig:NFNet_block_large}) followed by standard residual blocks (as shown in Figure~\ref{fig:NFNet_block_nontransition}). In all but the first stage, the transition block downsamples (with $2\times2$ average pooling on the skip path and by striding the first $3\times3$ convolution on the main path) and changes the output channel count (via a $1\times1$ shortcut convolution on the skip path). \citet{he2019bag} identified that the use of a $2\times2$ average pooling improves performance over using a strided $1\times1$ convolution on the skip path (which merely subsamples the activation). Note that this is slightly different from \citet{bello2021lambdanetworks}, which uses a $3\times3$ average pooling kernel with stride 2.

All blocks employ the pre-activation ResNe(X)t bottleneck pattern with an added $3\times3$ grouped convolution inside the bottleneck. This means that the main path comprises a $1\times1$ convolution whose output channels are equal to $0.5\times$ the output channel count for the block, two $3\times3$ grouped convolution with group width 128 (with the first strided in transition blocks), and a final $1\times1$ convolution whose output channel count is equal to the block output channel count.

Following the last $1\times1$ convolution is a Squeeze \& Excite layer \citep{hu2018squeeze}, which globally average pools the activation, applies two linear layers with an interleaved scaled nonlinearity to the pooled activation, applies a sigmoid, then rescales the tensor channel-wise by twice the value of this sigmoid. Concretely, the output of this layer is $2\sigma(FC(GELU(FC(pool(h))))) \times h$. The non-standard scalar multiplier of 2 is used following \citet{brock2021characterizing} to maintain signal variance.

After all of the residual stages, we apply a $1\times1$ expansion convolution that doubles the channel count, similar to the final expansion convolution in EfficientNets \citep{tan2019efficientnet}, then global average pooling. This layer is primarily helpful when using very thin networks, as it is typically desirable to have the dimensionality of the final activation vectors (which the classifier layer receives) be greater than or equal to the number of classes, but we retain it in our wider networks to benefit future work which might seek to train very thin networks based on our backbones. We tried replacing this convolution with a fully connected layer after the average pooling but found that this was not helpful. 

The final layer is a fully-connected classifier layer with learnable biases which outputs a 1000-way class vector (which can be softmaxed in order to obtain normalized class probabilities). We initialize this layer's weight with a standard deviation of $0.01$ following \citet{goyal2017accurate}. We found that initializing the weight with zeros as is sometimes done could sometimes lead to instabilities when training with very large numbers of output classes.

No activation normalization layers are used anywhere in our residual blocks. Instead, we employ the Normalizer-Free variance downscaling strategy \citep{brock2021characterizing}. This means that the input to the main path of the residual block is multiplied by $1/\beta$, where $\beta$ is the analytically predicted value of the variance at that block at initialization, and the output of the block is multiplied by a scalar hyperparameter $\alpha$, typically set to a small value like $\alpha=0.2$. 
As in \citet{brock2021characterizing}, we compute the expected empirical variance at residual block $\ell$ analytically using $\Var(x_\ell) = \Var(x_{\ell-1}) + \alpha^{2}$, with $\Var(x_{0}) = 1$, resulting in $\beta_\ell = \sqrt{\Var(x_\ell)}$. We also mimic the variance reset that happens in the transition blocks of batch-normalized networks, by having the shortcut convolution in transition layers operate on $(x_{\ell}/\beta_{\ell})$ rather than $x_{\ell}$ (see Figure~\ref{fig:NFNet_block_large}). This ensures unit signal variance at the start of each stage ($\Var(x_{\ell+1}) = 1 + \alpha^2$).

Additionally following \citet{brock2021characterizing}, we also employ SkipInit \citep{de2020batch}, a learnable zero-initialized scalar gain in addition to $\alpha$ which results in the residual block being initialized to the identity (except in transition layers), similar to \citet{goyal2017accurate, zhang2019fixup, bachlechner2020rezero}, which we find to improve stability for very deep networks.  While this will result in the signal propagation at initialization not actually following the expected variance as computed above, we find that the variance downscaling and $\alpha$ scalar are still beneficial for stability.

All convolutions employ Scaled Weight Standardization \citep{brock2021characterizing}, with a learnable affine gain applied to the standardized weight and a learnable affine bias applied to the output of the convolution operation. Critically, weight decay is not applied to the affine gains or biases or the SkipInit gains. The S\&E layers do not employ weight standardization on their fully connected layers, nor does the fully-connected classifier layer's weight. We initialize the underlying weights for these layers using LeCun initialization \citep{lecun2012efficient}.

%% file: tables/model_ablation_table_detailed.tex
\begin{center}\begin{tabular}{l| c c | c c | c c | c c }
\toprule [0.15em]
 & \multicolumn{2}{c|}{F0} & \multicolumn{2}{c|}{F1} & \multicolumn{2}{c|}{F2} & \multicolumn{2}{c|}{F3} \\
\midrule [0.1em]
    Baseline & 80.4\% & $58.0$ms & 81.7\% & $116.0$ms & 82.0\% & $211.7$ms & 82.3\% & $369.5$ms \\
    + Modified Width & 80.9\% & $64.1$ms & 81.8\% & $133.9$ms & 82.0\% & $252.2$ms & 82.3\% & $441.5$ms \\
    + Second Conv & 81.3\% & $73.3$ms & 82.2\% & $158.5$ms & 82.4\% & $295.8$ms & 82.7\% & $532.2$ms \\
    + MixUp & 82.2\% & $73.3$ms & 82.9\% & $158.5$ms & 83.1\% & $295.8$ms & 83.5\% & $532.2$ms \\
    + RandAugment & 83.2\% & $73.3$ms & 84.6\% & $158.5$ms & 84.8\% & $295.8$ms & 85.0\% & $532.2$ms \\
    + CutMix & 83.6\% & $73.3$ms & 84.7\% & $158.5$ms & 85.1\% & $295.8$ms & 85.7\% & $532.2$ms \\
    Default Width + Augs & 83.1\% & $65.9$ms & 84.5\% & $137.4$ms & 85.0\% & $248.8$ms & 85.5\% & $452.2$ms \\
    -NF, + BN & 83.4\% & $111.7$ms & 84.4\% & $258.0$ms & 85.1\% & $396.3$ms & 85.5\% & $617.7$ms \\
\bottomrule[0.15em]
\end{tabular}
\end{center}

%% file: Appendix_AGC_Ablations.tex
\begin{figure}[t]
	\includegraphics[width=0.68\columnwidth]
	{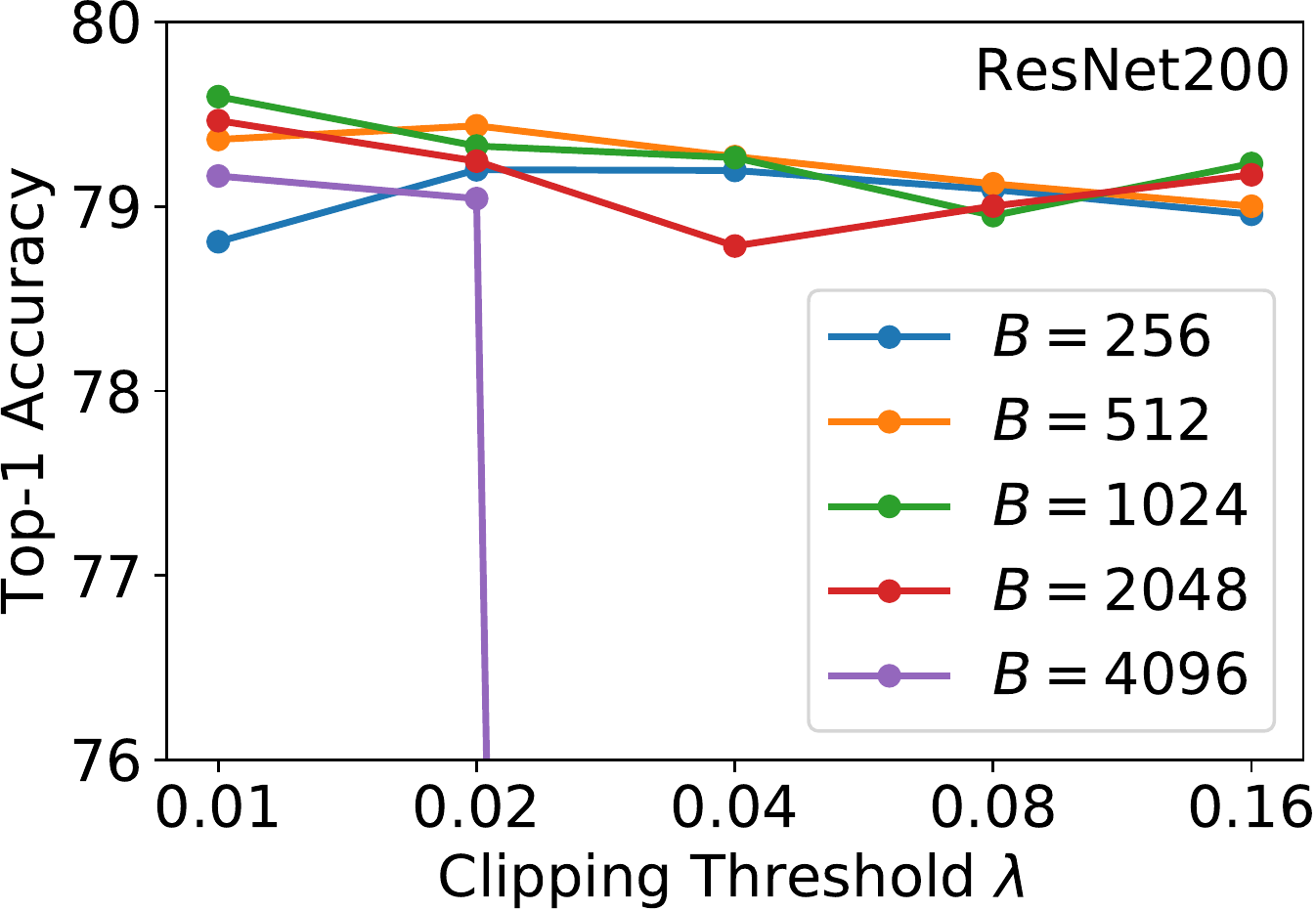}
\caption{Performance across different clipping thresholds $\lambda$ of AGC for different batch sizes on ResNet200.}
\label{fig:agc_ablations_clipping_ResNet200}
\end{figure}

In Figure \ref{fig:agc_ablations_clipping_ResNet200}, we show performance for different clipping thresholds $\lambda$ across a range of batch sizes on ResNet200, using the same training setup described in Section \ref{subsec:agc_ablations}. In both Figure \ref{fig:agc_ablations_clipping_ResNet200} and Figure \ref{fig:agc_batchsize_ablations}, we run NF-ResNets with AGC for 5 independent runs, and report the average of the best 4 of these 5 runs. This ensures that our results are robust to outliers and failed training runs.

As in Figure \ref{fig:agc_clippingsweep}, we see that smaller clipping thresholds are necessary for stability at higher batch sizes on the ResNet200. For all our experiments in Section \ref{sec:experiments} where we use batch size 4096, we use a clipping threshold $\lambda = 0.01$.

%% file: Appendix_Negative_Results.tex
\begin{figure}[htbp]
	\includegraphics[width=0.99\columnwidth]
	{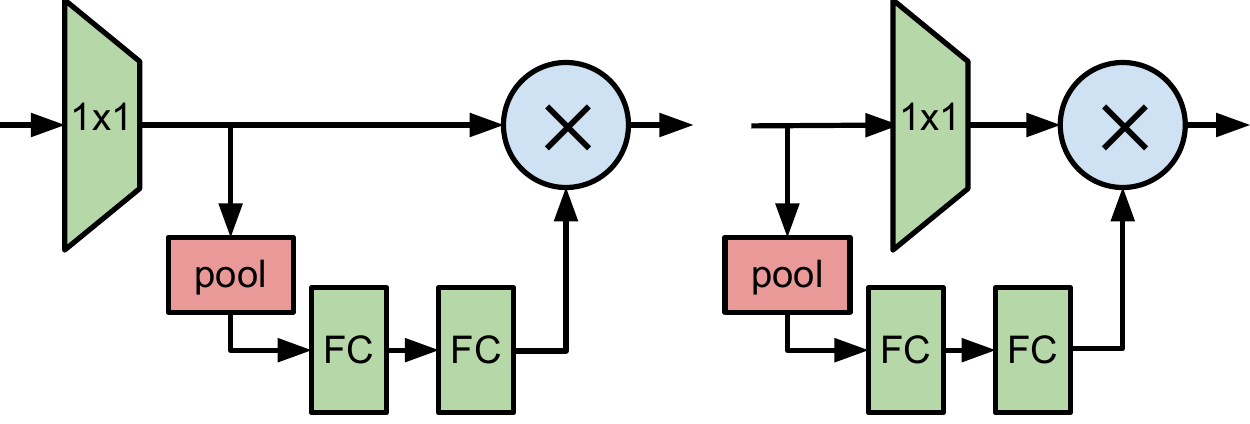}
	\caption{\label{fig:s_e_blocks} Comparison of standard (left) and ``straddling'' (right) Squeeze \& Excite blocks. Both forms of S\&E block allow for full cross-channel connectivity, but only in the form of a scalar multiplier per channel.} 
\end{figure}

In the course of developing the NFNet architecture we experimented with strategies impacting a range of model design aspects, including rules for picking backbone width and depth, bottleneck compression or expansion ratios, choice of group width, the placement of Squeeze \& Excite (S\&E) layers, and more. In this section we present select insights from what we found \textit{not} to work well. As in Section~\ref{sec:modeldesign}, our goal here was to improve the pareto front of top-1 holdout accuracy versus training speed.

First, we considered patterns where, for a given choice of backbone, we allowed the group width or number of groups in the $3\times3$ convolutions to be different in different stages, or similarly allowed the bottleneck ratio to vary in different stages. We also considered varying whether the transition blocks would have their bottleneck ratios be a function of the number of block output channels (as in ResNet models) or as the number of block input channels (as in many mobile models). For example, one model family variant used a group width of $[8, 8, 16, 16]$ in each of the four stages, with a bottleneck ratio of 0.25 (with the transition blocks using the bottleneck width based on the input channel count) in the first two stages and 0.5 in the latter two stages (with the transition blocks here using the bottleneck width based on the output channel count).

While we occasionally found that some of this variance could be helpful (for example, using inverted bottleneck blocks in the first stage yielded occasional but inconsistent improvements), we broadly found such heterogeneity to be unnecessary, and to confound attempts to reason out interpretable design patterns. We expect that these design aspects could yield better models if incorporated into large-scale architecture search to get individual models at given compute budget targets, but to be less useful for manual design. Our final NFNet designs are largely homogenous with respect to these parameters, with only width and stage depth varying between stages, and ResNet style bottleneck widths (where the channel count of the $3\times3$ convolutions is the number of output channels times the bottleneck ratio).

We explored aggressive downsampling strategies, such as operating on $8\times8$ DCT coefficients as in \citet{gueguen2018faster} instead of using the standard ResNet stem. While this is an effective way to improve model speed, we found that any improvements in model speed came at the cost of model accuracy. This appears to hold true even when this downsampling is done with an invertible operation (e.g. an orthogonal strided transform like the DCT) such that no information is lost. This is arguably consistent with the observations in \citet{sandler2019non}, suggesting that model ``internal resolution'' is a more important quantity to consider in this respect, but we did not explore this direction in further detail.

We next considered trying to improve speed by making our $1\times1$ dense convolutions into grouped convolutions. This normally causes sharp performance degradation, as these layers are responsible for the flow of information across all channels (as the other convolutions are grouped), and removing their full connectivity substantially reduces model expressivity. To ameliorate this we considered applying straddled Squeeze \& Excite layers, where the input to the S\&E is the input to the convolution, but the output of the S\&E multiplies the output of the convolution. This is in constrast to the normal Squeeze \& Excite formulation, which simply operates directly on an activation (i.e. it takes in a value, and its output is used to multiply that same value). Both forms of S\&E block help to restore cross-channel connectivity (albeit not as strongly as using a fully connected convolution) more cheaply than fully-connected layers as they operate on globally average-pooled activations.

Employing grouped $1\times1$ convs with a small number of groups (2 or 4) paired with S\&E layers slightly reduces accuracy, and improves theoretical FLOPS and reduces parameter counts, with the straddled S\&E block resulting in slightly improved accuracy relative to the standard S\&E block. However, we found there was no choice of $1\times1$ group width or group count which maintained comparable accuracy while reducing training latency. Using a high group count (and therefore a small group width) substantially reduces FLOPS and parameter counts, but also substantially reduces model performance, indicating that incorporating these S\&E layers helps but does not fully recover the expressivity of dense $1\times1$ convolutions.

Finally, we did not experiment with any attention variants \citep{bello2021lambdanetworks, srinivas2021bottleneck}, and we expect that our results could likely be improved by adopting these strategies into our models.

%% file: main.bbl
\begin{thebibliography}{92}
\providecommand{\natexlab}[1]{#1}
\providecommand{\url}[1]{\texttt{#1}}
\expandafter\ifx\csname urlstyle\endcsname\relax
  \providecommand{\doi}[1]{doi: #1}\else
  \providecommand{\doi}{doi: \begingroup \urlstyle{rm}\Url}\fi

\bibitem[Arpit et~al.(2016)Arpit, Zhou, Kota, and
  Govindaraju]{arpit2016normalization}
Arpit, D., Zhou, Y., Kota, B., and Govindaraju, V.
\newblock Normalization propagation: A parametric technique for removing
  internal covariate shift in deep networks.
\newblock In \emph{International Conference on Machine Learning}, pp.\
  1168--1176, 2016.

\bibitem[Ba et~al.(2016)Ba, Kiros, and Hinton]{ba2016layer}
Ba, J.~L., Kiros, J.~R., and Hinton, G.~E.
\newblock Layer normalization.
\newblock \emph{arXiv preprint arXiv:1607.06450}, 2016.

\bibitem[Babuschkin et~al.(2020)Babuschkin, Baumli, Bell, Bhupatiraju, Bruce,
  Buchlovsky, Budden, Cai, Clark, Danihelka, Fantacci, Godwin, Jones, Hennigan,
  Hessel, Kapturowski, Keck, Kemaev, King, Martens, Mikulik, Norman, Quan,
  Papamakarios, Ring, Ruiz, Sanchez, Schneider, Sezener, Spencer, Srinivasan,
  Stokowiec, and Viola]{deepmind2020jax}
Babuschkin, I., Baumli, K., Bell, A., Bhupatiraju, S., Bruce, J., Buchlovsky,
  P., Budden, D., Cai, T., Clark, A., Danihelka, I., Fantacci, C., Godwin, J.,
  Jones, C., Hennigan, T., Hessel, M., Kapturowski, S., Keck, T., Kemaev, I.,
  King, M., Martens, L., Mikulik, V., Norman, T., Quan, J., Papamakarios, G.,
  Ring, R., Ruiz, F., Sanchez, A., Schneider, R., Sezener, E., Spencer, S.,
  Srinivasan, S., Stokowiec, W., and Viola, F.
\newblock The {D}eep{M}ind {JAX} {E}cosystem, 2020.
\newblock URL \url{http://github.com/deepmind}.

\bibitem[Bachlechner et~al.(2020)Bachlechner, Majumder, Mao, Cottrell, and
  McAuley]{bachlechner2020rezero}
Bachlechner, T., Majumder, B.~P., Mao, H.~H., Cottrell, G.~W., and McAuley, J.
\newblock Rezero is all you need: Fast convergence at large depth.
\newblock \emph{arXiv preprint arXiv:2003.04887}, 2020.

\bibitem[Balduzzi et~al.(2017)Balduzzi, Frean, Leary, Lewis, Ma, and
  McWilliams]{balduzzi2017shattered}
Balduzzi, D., Frean, M., Leary, L., Lewis, J., Ma, K. W.-D., and McWilliams, B.
\newblock The shattered gradients problem: If resnets are the answer, then what
  is the question?
\newblock In \emph{International Conference on Machine Learning}, pp.\
  342--350, 2017.

\bibitem[Bello(2021)]{bello2021lambdanetworks}
Bello, I.
\newblock Lambdanetworks: Modeling long-range interactions without attention.
\newblock In \emph{International Conference on Learning Representations
  {ICLR}}, 2021.
\newblock URL \url{https://openreview.net/forum?id=xTJEN-ggl1b}.

\bibitem[Bernstein et~al.(2020)Bernstein, Vahdat, Yue, and
  Liu]{bernstein2020distance}
Bernstein, J., Vahdat, A., Yue, Y., and Liu, M.-Y.
\newblock On the distance between two neural networks and the stability of
  learning.
\newblock \emph{arXiv preprint arXiv:2002.03432}, 2020.

\bibitem[Bjorck et~al.(2018)Bjorck, Gomes, Selman, and
  Weinberger]{bjorck2018understanding}
Bjorck, N., Gomes, C.~P., Selman, B., and Weinberger, K.~Q.
\newblock Understanding batch normalization.
\newblock In \emph{Advances in Neural Information Processing Systems}, pp.\
  7694--7705, 2018.

\bibitem[Bradbury et~al.(2018)Bradbury, Frostig, Hawkins, Johnson, Leary,
  Maclaurin, and Wanderman-Milne]{jax2018github}
Bradbury, J., Frostig, R., Hawkins, P., Johnson, M.~J., Leary, C., Maclaurin,
  D., and Wanderman-Milne, S.
\newblock {JAX}: composable transformations of {P}ython+{N}um{P}y programs,
  2018.
\newblock URL \url{http://github.com/google/jax}.

\bibitem[Brock et~al.(2021)Brock, De, and Smith]{brock2021characterizing}
Brock, A., De, S., and Smith, S.~L.
\newblock Characterizing signal propagation to close the performance gap in
  unnormalized resnets.
\newblock In \emph{9th International Conference on Learning Representations,
  {ICLR}}, 2021.

\bibitem[Chen et~al.(2020)Chen, Kornblith, Norouzi, and Hinton]{chen2020simple}
Chen, T., Kornblith, S., Norouzi, M., and Hinton, G.
\newblock A simple framework for contrastive learning of visual
  representations.
\newblock In \emph{International conference on machine learning}, pp.\
  1597--1607. PMLR, 2020.

\bibitem[Cubuk et~al.(2020)Cubuk, Zoph, Shlens, and Le]{cubuk2020randaugment}
Cubuk, E.~D., Zoph, B., Shlens, J., and Le, Q.~V.
\newblock Randaugment: Practical automated data augmentation with a reduced
  search space.
\newblock In \emph{Proceedings of the IEEE/CVF Conference on Computer Vision
  and Pattern Recognition Workshops}, pp.\  702--703, 2020.

\bibitem[De \& Smith(2020)De and Smith]{de2020batch}
De, S. and Smith, S.
\newblock Batch normalization biases residual blocks towards the identity
  function in deep networks.
\newblock \emph{Advances in Neural Information Processing Systems}, 33, 2020.

\bibitem[Dosovitskiy et~al.(2021)Dosovitskiy, Beyer, Kolesnikov, Weissenborn,
  Zhai, Unterthiner, Dehghani, Minderer, Heigold, Gelly, Uszkoreit, and
  Houlsby]{dosovitskiy2021an}
Dosovitskiy, A., Beyer, L., Kolesnikov, A., Weissenborn, D., Zhai, X.,
  Unterthiner, T., Dehghani, M., Minderer, M., Heigold, G., Gelly, S.,
  Uszkoreit, J., and Houlsby, N.
\newblock An image is worth 16x16 words: Transformers for image recognition at
  scale.
\newblock In \emph{9th International Conference on Learning Representations,
  {ICLR}}, 2021.
\newblock URL \url{https://openreview.net/forum?id=YicbFdNTTy}.

\bibitem[Foret et~al.(2021)Foret, Kleiner, Mobahi, and
  Neyshabur]{foret2021sharpnessaware}
Foret, P., Kleiner, A., Mobahi, H., and Neyshabur, B.
\newblock Sharpness-aware minimization for efficiently improving
  generalization.
\newblock In \emph{9th International Conference on Learning Representations,
  {ICLR}}, 2021.
\newblock URL \url{https://openreview.net/forum?id=6Tm1mposlrM}.

\bibitem[Gitman \& Ginsburg(2017)Gitman and Ginsburg]{gitman2017comparison}
Gitman, I. and Ginsburg, B.
\newblock Comparison of batch normalization and weight normalization algorithms
  for the large-scale image classification.
\newblock \emph{arXiv preprint arXiv:1709.08145}, 2017.

\bibitem[Gong et~al.(2020)Gong, Ren, Ye, and Liu]{gong2020maxup}
Gong, C., Ren, T., Ye, M., and Liu, Q.
\newblock Maxup: A simple way to improve generalization of neural network
  training.
\newblock \emph{arXiv preprint arXiv:2002.09024}, 2020.

\bibitem[Google(2021)]{tpu_performance}
Google.
\newblock {Cloud TPU Performance Guide}.
\newblock \url{https://cloud.google.com/tpu/docs/performance-guide}, 2021.

\bibitem[Goyal et~al.(2017)Goyal, Doll{\'a}r, Girshick, Noordhuis, Wesolowski,
  Kyrola, Tulloch, Jia, and He]{goyal2017accurate}
Goyal, P., Doll{\'a}r, P., Girshick, R., Noordhuis, P., Wesolowski, L., Kyrola,
  A., Tulloch, A., Jia, Y., and He, K.
\newblock Accurate, large minibatch sgd: Training imagenet in 1 hour.
\newblock \emph{arXiv preprint arXiv:1706.02677}, 2017.

\bibitem[Gueguen et~al.(2018)Gueguen, Sergeev, Kadlec, Liu, and
  Yosinski]{gueguen2018faster}
Gueguen, L., Sergeev, A., Kadlec, B., Liu, R., and Yosinski, J.
\newblock Faster neural networks straight from jpeg.
\newblock \emph{Advances in Neural Information Processing Systems},
  31:\penalty0 3933--3944, 2018.

\bibitem[Hanin \& Rolnick(2018)Hanin and Rolnick]{hanin2018start}
Hanin, B. and Rolnick, D.
\newblock How to start training: The effect of initialization and architecture.
\newblock In \emph{Advances in Neural Information Processing Systems}, pp.\
  571--581, 2018.

\bibitem[Harris et~al.(2020)Harris, Millman, van~der Walt, Gommers, Virtanen,
  Cournapeau, Wieser, Taylor, Berg, Smith, Kern, Picus, Hoyer, van Kerkwijk,
  Brett, Haldane, del R{\'i}o, Wiebe, Peterson, G{\'e}rard-Marchant, Sheppard,
  Reddy, Weckesser, Abbasi, Gohlke, and Oliphant]{harris2020numpy}
Harris, C.~R., Millman, K.~J., van~der Walt, S.~J., Gommers, R., Virtanen, P.,
  Cournapeau, D., Wieser, E., Taylor, J., Berg, S., Smith, N.~J., Kern, R.,
  Picus, M., Hoyer, S., van Kerkwijk, M.~H., Brett, M., Haldane, A., del
  R{\'i}o, J.~F., Wiebe, M., Peterson, P., G{\'e}rard-Marchant, P., Sheppard,
  K., Reddy, T., Weckesser, W., Abbasi, H., Gohlke, C., and Oliphant, T.~E.
\newblock Array programming with numpy.
\newblock \emph{Nature}, 585\penalty0 (7825):\penalty0 357--362, Sep 2020.
\newblock ISSN 1476-4687.

\bibitem[He et~al.(2016{\natexlab{a}})He, Zhang, Ren, and Sun]{he2016identity}
He, K., Zhang, X., Ren, S., and Sun, J.
\newblock Identity mappings in deep residual networks.
\newblock In \emph{European conference on computer vision}, pp.\  630--645.
  Springer, 2016{\natexlab{a}}.

\bibitem[He et~al.(2016{\natexlab{b}})He, Zhang, Ren, and Sun]{he2016resnets}
He, K., Zhang, X., Ren, S., and Sun, J.
\newblock Deep residual learning for image recognition.
\newblock In \emph{CVPR}, 2016{\natexlab{b}}.

\bibitem[He et~al.(2020)He, Fan, Wu, Xie, and Girshick]{he2020momentum}
He, K., Fan, H., Wu, Y., Xie, S., and Girshick, R.
\newblock Momentum contrast for unsupervised visual representation learning.
\newblock In \emph{Proceedings of the IEEE/CVF Conference on Computer Vision
  and Pattern Recognition}, pp.\  9729--9738, 2020.

\bibitem[He et~al.(2019)He, Zhang, Zhang, Zhang, Xie, and Li]{he2019bag}
He, T., Zhang, Z., Zhang, H., Zhang, Z., Xie, J., and Li, M.
\newblock Bag of tricks for image classification with convolutional neural
  networks.
\newblock In \emph{Proceedings of the IEEE Conference on Computer Vision and
  Pattern Recognition}, pp.\  558--567, 2019.

\bibitem[Hendrycks \& Gimpel(2016)Hendrycks and Gimpel]{hendrycks2016gaussian}
Hendrycks, D. and Gimpel, K.
\newblock Gaussian error linear units ({GELUs}).
\newblock \emph{arXiv preprint arXiv:1606.08415}, 2016.

\bibitem[Hennigan et~al.(2020)Hennigan, Cai, Norman, and
  Babuschkin]{haiku2020github}
Hennigan, T., Cai, T., Norman, T., and Babuschkin, I.
\newblock {H}aiku: {S}onnet for {JAX}, 2020.
\newblock URL \url{http://github.com/deepmind/dm-haiku}.

\bibitem[Hoffer et~al.(2017)Hoffer, Hubara, and Soudry]{hoffer2017train}
Hoffer, E., Hubara, I., and Soudry, D.
\newblock Train longer, generalize better: closing the generalization gap in
  large batch training of neural networks.
\newblock In \emph{Advances in Neural Information Processing Systems}, pp.\
  1731--1741, 2017.

\bibitem[Hooker(2020)]{hooker2020hardware}
Hooker, S.
\newblock The hardware lottery.
\newblock \emph{arXiv preprint arXiv:2009.06489}, 2020.

\bibitem[Hu et~al.(2018)Hu, Shen, and Sun]{hu2018squeeze}
Hu, J., Shen, L., and Sun, G.
\newblock Squeeze-and-excitation networks.
\newblock In \emph{Proceedings of the IEEE conference on computer vision and
  pattern recognition}, pp.\  7132--7141, 2018.

\bibitem[Huang et~al.(2016)Huang, Sun, Liu, Sedra, and
  Weinberger]{huang2016deep}
Huang, G., Sun, Y., Liu, Z., Sedra, D., and Weinberger, K.~Q.
\newblock Deep networks with stochastic depth.
\newblock In \emph{European conference on computer vision}, pp.\  646--661.
  Springer, 2016.

\bibitem[Huang et~al.(2017)Huang, Liu, Liu, Lang, and Tao]{huang2017centered}
Huang, L., Liu, X., Liu, Y., Lang, B., and Tao, D.
\newblock Centered weight normalization in accelerating training of deep neural
  networks.
\newblock In \emph{Proceedings of the IEEE International Conference on Computer
  Vision}, pp.\  2803--2811, 2017.

\bibitem[Huang et~al.(2020)Huang, Qin, Zhou, Zhu, Liu, and
  Shao]{huang2020normalization}
Huang, L., Qin, J., Zhou, Y., Zhu, F., Liu, L., and Shao, L.
\newblock Normalization techniques in training dnns: Methodology, analysis and
  application.
\newblock \emph{arXiv preprint arXiv:2009.12836}, 2020.

\bibitem[Ioffe(2017)]{ioffe2017batch}
Ioffe, S.
\newblock Batch renormalization: Towards reducing minibatch dependence in
  batch-normalized models.
\newblock \emph{arXiv preprint arXiv:1702.03275}, 2017.

\bibitem[Ioffe \& Szegedy(2015)Ioffe and Szegedy]{ioffe2015batchnorm}
Ioffe, S. and Szegedy, C.
\newblock Batch normalization: Accelerating deep network training by reducing
  internal covariate shift.
\newblock In \emph{ICML}, 2015.

\bibitem[Jacot et~al.(2019)Jacot, Gabriel, and Hongler]{jacot2019freeze}
Jacot, A., Gabriel, F., and Hongler, C.
\newblock Freeze and chaos for dnns: an ntk view of batch normalization,
  checkerboard and boundary effects.
\newblock \emph{arXiv preprint arXiv:1907.05715}, 2019.

\bibitem[Kaplan et~al.(2020)Kaplan, McCandlish, Henighan, Brown, Chess, Child,
  Gray, Radford, Wu, and Amodei]{kaplan2020scaling}
Kaplan, J., McCandlish, S., Henighan, T., Brown, T.~B., Chess, B., Child, R.,
  Gray, S., Radford, A., Wu, J., and Amodei, D.
\newblock Scaling laws for neural language models.
\newblock \emph{arXiv preprint arXiv:2001.08361}, 2020.

\bibitem[Kolesnikov et~al.(2019)Kolesnikov, Beyer, Zhai, Puigcerver, Yung,
  Gelly, and Houlsby]{kolesnikov2019large}
Kolesnikov, A., Beyer, L., Zhai, X., Puigcerver, J., Yung, J., Gelly, S., and
  Houlsby, N.
\newblock Large scale learning of general visual representations for transfer.
\newblock \emph{arXiv preprint arXiv:1912.11370}, 2019.

\bibitem[Krizhevsky et~al.(2012)Krizhevsky, Sutskever, and
  Hinton]{krizhevsky2012imagenet}
Krizhevsky, A., Sutskever, I., and Hinton, G.~E.
\newblock Imagenet classification with deep convolutional neural networks.
\newblock \emph{Advances in neural information processing systems},
  25:\penalty0 1097--1105, 2012.

\bibitem[LeCun et~al.(2012)LeCun, Bottou, Orr, and
  M{\"u}ller]{lecun2012efficient}
LeCun, Y.~A., Bottou, L., Orr, G.~B., and M{\"u}ller, K.-R.
\newblock Efficient backprop.
\newblock In \emph{Neural networks: Tricks of the trade}, pp.\  9--48.
  Springer, 2012.

\bibitem[Loshchilov \& Hutter(2016)Loshchilov and Hutter]{loshchilov2016sgdr}
Loshchilov, I. and Hutter, F.
\newblock Sgdr: Stochastic gradient descent with warm restarts.
\newblock \emph{arXiv preprint arXiv:1608.03983}, 2016.

\bibitem[Loshchilov \& Hutter(2017)Loshchilov and
  Hutter]{loshchilov2017decoupled}
Loshchilov, I. and Hutter, F.
\newblock Decoupled weight decay regularization.
\newblock \emph{arXiv preprint arXiv:1711.05101}, 2017.

\bibitem[Luo et~al.(2018)Luo, Wang, Shao, and Peng]{luo2018towards}
Luo, P., Wang, X., Shao, W., and Peng, Z.
\newblock Towards understanding regularization in batch normalization.
\newblock \emph{arXiv preprint arXiv:1809.00846}, 2018.

\bibitem[Mahajan et~al.(2018)Mahajan, Girshick, Ramanathan, He, Paluri, Li,
  Bharambe, and Van Der~Maaten]{mahajan2018exploring}
Mahajan, D., Girshick, R., Ramanathan, V., He, K., Paluri, M., Li, Y.,
  Bharambe, A., and Van Der~Maaten, L.
\newblock Exploring the limits of weakly supervised pretraining.
\newblock In \emph{Proceedings of the European Conference on Computer Vision
  {ECCV}}, pp.\  181--196, 2018.

\bibitem[Merity et~al.(2018)Merity, Keskar, and Socher]{merity2018regularizing}
Merity, S., Keskar, N.~S., and Socher, R.
\newblock Regularizing and optimizing {LSTM} language models.
\newblock In \emph{International Conference on Learning Representations}, 2018.

\bibitem[Nesterov(1983)]{nesterov1983}
Nesterov, Y.
\newblock A method for unconstrained convex minimization problem with the rate
  of convergence $ o(1/k^2)$.
\newblock \emph{Doklady AN USSR}, pp.\  (269), 543--547, 1983.

\bibitem[Pascanu et~al.(2013)Pascanu, Mikolov, and
  Bengio]{pascanu2013difficulty}
Pascanu, R., Mikolov, T., and Bengio, Y.
\newblock On the difficulty of training recurrent neural networks.
\newblock In \emph{International conference on machine learning}, pp.\
  1310--1318, 2013.

\bibitem[Pham et~al.(2020)Pham, Xie, Dai, and Le]{pham2020meta}
Pham, H., Xie, Q., Dai, Z., and Le, Q.~V.
\newblock Meta pseudo labels.
\newblock \emph{arXiv preprint arXiv:2003.10580}, 2020.

\bibitem[Pham et~al.(2019)Pham, Lutellier, Qi, and Tan]{pham2019cradle}
Pham, H.~V., Lutellier, T., Qi, W., and Tan, L.
\newblock Cradle: cross-backend validation to detect and localize bugs in deep
  learning libraries.
\newblock In \emph{2019 IEEE/ACM 41st International Conference on Software
  Engineering (ICSE)}, pp.\  1027--1038. IEEE, 2019.

\bibitem[Polyak(1964)]{polyak1964some}
Polyak, B.
\newblock Some methods of speeding up the convergence of iteration methods.
\newblock \emph{USSR Computational Mathematics and Mathematical Physics}, pp.\
  4(5):1--17, 1964.

\bibitem[Qiao et~al.(2019)Qiao, Wang, Liu, Shen, and Yuille]{qiao2019weight}
Qiao, S., Wang, H., Liu, C., Shen, W., and Yuille, A.
\newblock Weight standardization.
\newblock \emph{arXiv preprint arXiv:1903.10520}, 2019.

\bibitem[Qin et~al.(2020)Qin, Fang, Zhang, Liu, Wang, and
  Wang]{qin2020resizemix}
Qin, J., Fang, J., Zhang, Q., Liu, W., Wang, X., and Wang, X.
\newblock Resizemix: Mixing data with preserved object information and true
  labels.
\newblock \emph{arXiv preprint arXiv:2012.11101}, 2020.

\bibitem[Radford et~al.(2016)Radford, Metz, and Chintala]{radford2016dcgan}
Radford, A., Metz, L., and Chintala, S.
\newblock Unsupervised representation learning with deep convolutional
  generative adversarial networks.
\newblock In \emph{4th International Conference on Learning Representations,
  {ICLR}}, 2016.

\bibitem[Radosavovic et~al.(2020)Radosavovic, Kosaraju, Girshick, He, and
  Doll{\'a}r]{radosavovic2020designing}
Radosavovic, I., Kosaraju, R.~P., Girshick, R., He, K., and Doll{\'a}r, P.
\newblock Designing network design spaces.
\newblock In \emph{Proceedings of the IEEE/CVF Conference on Computer Vision
  and Pattern Recognition}, pp.\  10428--10436, 2020.

\bibitem[Raghu et~al.(2017{\natexlab{a}})Raghu, Gilmer, Yosinski, and
  Sohl-Dickstein]{raghu2017svcca}
Raghu, M., Gilmer, J., Yosinski, J., and Sohl-Dickstein, J.
\newblock Svcca: Singular vector canonical correlation analysis for deep
  learning dynamics and interpretability.
\newblock \emph{Advances in neural information processing systems},
  30:\penalty0 6076--6085, 2017{\natexlab{a}}.

\bibitem[Raghu et~al.(2017{\natexlab{b}})Raghu, Poole, Kleinberg, Ganguli, and
  Sohl-Dickstein]{raghu2017expressive}
Raghu, M., Poole, B., Kleinberg, J., Ganguli, S., and Sohl-Dickstein, J.
\newblock On the expressive power of deep neural networks.
\newblock In \emph{international conference on machine learning}, pp.\
  2847--2854. PMLR, 2017{\natexlab{b}}.

\bibitem[Robbins \& Monro(1951)Robbins and Monro]{robbins1951A}
Robbins, H. and Monro, S.
\newblock A stochastic approximation method.
\newblock \emph{The Annals of Mathematical Statistics}, pp.\  22(3):400--407,
  1951.

\bibitem[Rota~Bul{\`o} et~al.(2018)Rota~Bul{\`o}, Porzi, and
  Kontschieder]{rota2018place}
Rota~Bul{\`o}, S., Porzi, L., and Kontschieder, P.
\newblock In-place activated batchnorm for memory-optimized training of dnns.
\newblock In \emph{Proceedings of the IEEE Conference on Computer Vision and
  Pattern Recognition}, pp.\  5639--5647, 2018.

\bibitem[Russakovsky et~al.(2015)Russakovsky, Deng, Su, Krause, Satheesh, Ma,
  Huang, Karpathy, Khosla, Bernstein, Berg, and Fei-Fei]{ILSVRC2015}
Russakovsky, O., Deng, J., Su, H., Krause, J., Satheesh, S., Ma, S., Huang, Z.,
  Karpathy, A., Khosla, A., Bernstein, M., Berg, A.~C., and Fei-Fei, L.
\newblock Image{N}et large scale visual recognition challenge.
\newblock \emph{IJCV}, 115:\penalty0 211--252, 2015.

\bibitem[Sandler et~al.(2018)Sandler, Howard, Zhu, Zhmoginov, and
  Chen]{sandler2018mobilenetv2}
Sandler, M., Howard, A., Zhu, M., Zhmoginov, A., and Chen, L.-C.
\newblock Mobilenetv2: Inverted residuals and linear bottlenecks.
\newblock In \emph{Proceedings of the IEEE conference on computer vision and
  pattern recognition}, pp.\  4510--4520, 2018.

\bibitem[Sandler et~al.(2019)Sandler, Baccash, Zhmoginov, and
  Howard]{sandler2019non}
Sandler, M., Baccash, J., Zhmoginov, A., and Howard, A.
\newblock Non-discriminative data or weak model? on the relative importance of
  data and model resolution.
\newblock In \emph{Proceedings of the IEEE/CVF International Conference on
  Computer Vision Workshops}, pp.\  0--0, 2019.

\bibitem[Santurkar et~al.(2018)Santurkar, Tsipras, Ilyas, and
  Madry]{santurkar2018does}
Santurkar, S., Tsipras, D., Ilyas, A., and Madry, A.
\newblock How does batch normalization help optimization?
\newblock In \emph{Advances in Neural Information Processing Systems}, pp.\
  2483--2493, 2018.

\bibitem[Shao et~al.(2020)Shao, Hu, Wang, Xue, and Raj]{shao2020normalization}
Shao, J., Hu, K., Wang, C., Xue, X., and Raj, B.
\newblock Is normalization indispensable for training deep neural network?
\newblock \emph{Advances in Neural Information Processing Systems}, 33, 2020.

\bibitem[Shen et~al.(2020)Shen, Yao, Gholami, Mahoney, and
  Keutzer]{shen2020powernorm}
Shen, S., Yao, Z., Gholami, A., Mahoney, M., and Keutzer, K.
\newblock Powernorm: Rethinking batch normalization in transformers.
\newblock In \emph{International Conference on Machine Learning}, pp.\
  8741--8751. PMLR, 2020.

\bibitem[Simonyan \& Zisserman(2015)Simonyan and Zisserman]{simonyan2015vgg}
Simonyan, K. and Zisserman, A.
\newblock Very deep convolutional networks for large-scale image recognition.
\newblock In \emph{3rd International Conference on Learning Representations,
  {ICLR}}, 2015.

\bibitem[Singh \& Shrivastava(2019)Singh and Shrivastava]{singh2019evalnorm}
Singh, S. and Shrivastava, A.
\newblock Evalnorm: Estimating batch normalization statistics for evaluation.
\newblock In \emph{Proceedings of the IEEE/CVF International Conference on
  Computer Vision}, pp.\  3633--3641, 2019.

\bibitem[Smith et~al.(2020)Smith, Elsen, and De]{smith2020generalization}
Smith, S., Elsen, E., and De, S.
\newblock On the generalization benefit of noise in stochastic gradient
  descent.
\newblock In \emph{International Conference on Machine Learning}, pp.\
  9058--9067. PMLR, 2020.

\bibitem[Srinivas et~al.(2021)Srinivas, Lin, Parmar, Shlens, Abbeel, and
  Vaswani]{srinivas2021bottleneck}
Srinivas, A., Lin, T.-Y., Parmar, N., Shlens, J., Abbeel, P., and Vaswani, A.
\newblock Bottleneck transformers for visual recognition.
\newblock \emph{arXiv preprint arXiv:2101.11605}, 2021.

\bibitem[Srivastava et~al.(2014)Srivastava, Hinton, Krizhevsky, Sutskever, and
  Salakhutdinov]{srivastava2014dropout}
Srivastava, N., Hinton, G., Krizhevsky, A., Sutskever, I., and Salakhutdinov,
  R.
\newblock Dropout: a simple way to prevent neural networks from overfitting.
\newblock \emph{The Journal of Machine Learning Research}, 15\penalty0
  (1):\penalty0 1929--1958, 2014.

\bibitem[Srivastava et~al.(2015)Srivastava, Greff, and
  Schmidhuber]{srivastava2015highway}
Srivastava, R.~K., Greff, K., and Schmidhuber, J.
\newblock Highway networks.
\newblock \emph{arXiv preprint arXiv:1505.00387}, 2015.

\bibitem[Summers \& Dinneen(2019)Summers and Dinneen]{summers2019four}
Summers, C. and Dinneen, M.~J.
\newblock Four things everyone should know to improve batch normalization.
\newblock \emph{arXiv preprint arXiv:1906.03548}, 2019.

\bibitem[Sun et~al.(2017)Sun, Shrivastava, Singh, and Gupta]{sun17revisiting}
Sun, C., Shrivastava, A., Singh, S., and Gupta, A.
\newblock Revisiting unreasonable effectiveness of data in deep learning era.
\newblock In \emph{ICCV}, 2017.

\bibitem[Sutskever et~al.(2013)Sutskever, Martens, Dahl, and
  Hinton]{sutskever2013importance}
Sutskever, I., Martens, J., Dahl, G., and Hinton, G.
\newblock On the importance of initialization and momentum in deep learning.
\newblock In \emph{International conference on machine learning}, pp.\
  1139--1147, 2013.

\bibitem[Szegedy et~al.(2016{\natexlab{a}})Szegedy, Ioffe, Vanhoucke, and
  Alemi]{szegedy2016inception}
Szegedy, C., Ioffe, S., Vanhoucke, V., and Alemi, A.
\newblock Inception-v4, inception-resnet and the impact of residual connections
  on learning.
\newblock \emph{arXiv preprint arXiv:1602.07261}, 2016{\natexlab{a}}.

\bibitem[Szegedy et~al.(2016{\natexlab{b}})Szegedy, Vanhoucke, Ioffe, Shlens,
  and Wojna]{szegedy2016rethinking}
Szegedy, C., Vanhoucke, V., Ioffe, S., Shlens, J., and Wojna, Z.
\newblock Rethinking the inception architecture for computer vision.
\newblock In \emph{2016 IEEE Conference on Computer Vision and Pattern
  Recognition (CVPR)}, pp.\  2818--2826, 2016{\natexlab{b}}.

\bibitem[Tan \& Le(2019)Tan and Le]{tan2019efficientnet}
Tan, M. and Le, Q.
\newblock Efficientnet: Rethinking model scaling for convolutional neural
  networks.
\newblock In \emph{International Conference on Machine Learning}, pp.\
  6105--6114, 2019.

\bibitem[Tieleman \& Hinton(2012)Tieleman and Hinton]{tieleman2012rmsprop}
Tieleman, T. and Hinton, G.
\newblock Rmsprop: Divide the gradient by a running average of its recent
  magnitude.
\newblock \emph{COURSERA: Neural networks for machine learning}, pp.\
  4(2):26--31, 2012.

\bibitem[Touvron et~al.(2019)Touvron, Vedaldi, Douze, and
  J{\'e}gou]{touvron2019fixing}
Touvron, H., Vedaldi, A., Douze, M., and J{\'e}gou, H.
\newblock Fixing the train-test resolution discrepancy.
\newblock In \emph{Advances in Neural Information Processing Systems}, pp.\
  8252--8262, 2019.

\bibitem[Touvron et~al.(2020)Touvron, Cord, Douze, Massa, Sablayrolles, and
  J{\'e}gou]{touvron2020training}
Touvron, H., Cord, M., Douze, M., Massa, F., Sablayrolles, A., and J{\'e}gou,
  H.
\newblock Training data-efficient image transformers \& distillation through
  attention.
\newblock \emph{arXiv preprint arXiv:2012.12877}, 2020.

\bibitem[Vaswani et~al.(2017)Vaswani, Shazeer, Parmar, Uszkoreit, Jones, Gomez,
  Kaiser, and Polosukhin]{vaswani2017attention}
Vaswani, A., Shazeer, N., Parmar, N., Uszkoreit, J., Jones, L., Gomez, A.~N.,
  Kaiser, L., and Polosukhin, I.
\newblock Attention is all you need.
\newblock \emph{arXiv preprint arXiv:1706.03762}, 2017.

\bibitem[Wu \& He(2018)Wu and He]{wu2018group}
Wu, Y. and He, K.
\newblock Group normalization.
\newblock In \emph{Proceedings of the European Conference on Computer Vision
  (ECCV)}, pp.\  3--19, 2018.

\bibitem[Xie et~al.(2020)Xie, Luong, Hovy, and Le]{xie2020self}
Xie, Q., Luong, M.-T., Hovy, E., and Le, Q.~V.
\newblock Self-training with noisy student improves imagenet classification.
\newblock In \emph{Proceedings of the IEEE/CVF Conference on Computer Vision
  and Pattern Recognition}, pp.\  10687--10698, 2020.

\bibitem[Xie et~al.(2017)Xie, Girshick, Doll{\'a}r, Tu, and
  He]{xie2017aggregated}
Xie, S., Girshick, R., Doll{\'a}r, P., Tu, Z., and He, K.
\newblock Aggregated residual transformations for deep neural networks.
\newblock In \emph{Proceedings of the IEEE conference on computer vision and
  pattern recognition}, pp.\  1492--1500, 2017.

\bibitem[Yang et~al.(2019)Yang, Pennington, Rao, Sohl-Dickstein, and
  Schoenholz]{yang2019mean}
Yang, G., Pennington, J., Rao, V., Sohl-Dickstein, J., and Schoenholz, S.~S.
\newblock A mean field theory of batch normalization.
\newblock \emph{arXiv preprint arXiv:1902.08129}, 2019.

\bibitem[You et~al.(2017)You, Gitman, and Ginsburg]{you2017large}
You, Y., Gitman, I., and Ginsburg, B.
\newblock Large batch training of convolutional networks.
\newblock \emph{arXiv preprint arXiv:1708.03888}, 2017.

\bibitem[You et~al.(2019)You, Li, Reddi, Hseu, Kumar, Bhojanapalli, Song,
  Demmel, Keutzer, and Hsieh]{you2019large}
You, Y., Li, J., Reddi, S., Hseu, J., Kumar, S., Bhojanapalli, S., Song, X.,
  Demmel, J., Keutzer, K., and Hsieh, C.-J.
\newblock Large batch optimization for deep learning: Training bert in 76
  minutes.
\newblock In \emph{7th International Conference on Learning Representations,
  {ICLR}}, 2019.

\bibitem[Yun et~al.(2019)Yun, Han, Oh, Chun, Choe, and Yoo]{yun2019cutmix}
Yun, S., Han, D., Oh, S.~J., Chun, S., Choe, J., and Yoo, Y.
\newblock Cutmix: Regularization strategy to train strong classifiers with
  localizable features.
\newblock In \emph{Proceedings of the IEEE International Conference on Computer
  Vision}, pp.\  6023--6032, 2019.

\bibitem[Zhang et~al.(2017)Zhang, Cisse, Dauphin, and
  Lopez-Paz]{zhang2017mixup}
Zhang, H., Cisse, M., Dauphin, Y.~N., and Lopez-Paz, D.
\newblock mixup: Beyond empirical risk minimization.
\newblock \emph{arXiv preprint arXiv:1710.09412}, 2017.

\bibitem[Zhang et~al.(2019{\natexlab{a}})Zhang, Dauphin, and
  Ma]{zhang2019fixup}
Zhang, H., Dauphin, Y.~N., and Ma, T.
\newblock Fixup initialization: Residual learning without normalization.
\newblock \emph{arXiv preprint arXiv:1901.09321}, 2019{\natexlab{a}}.

\bibitem[Zhang et~al.(2019{\natexlab{b}})Zhang, Goodfellow, Metaxas, and
  Odena]{zhang2019self}
Zhang, H., Goodfellow, I., Metaxas, D., and Odena, A.
\newblock Self-attention generative adversarial networks.
\newblock In \emph{International conference on machine learning}, pp.\
  7354--7363. PMLR, 2019{\natexlab{b}}.

\bibitem[Zhang et~al.(2020)Zhang, He, Sra, and Jadbabaie]{zhang2019gradient}
Zhang, J., He, T., Sra, S., and Jadbabaie, A.
\newblock Why gradient clipping accelerates training: A theoretical
  justification for adaptivity.
\newblock In \emph{8th International Conference on Learning Representations,
  {ICLR}}, 2020.
\newblock URL \url{https://openreview.net/forum?id=BJgnXpVYwS}.

\end{thebibliography}
